\documentclass[twoside,11pt]{article}

\usepackage[abbrvbib, preprint]{jmlr2e}

\usepackage[table, svgnames, dvipsnames]{xcolor}
\usepackage[utf8]{inputenc} 
\usepackage{hyperref}       
\usepackage{natbib}
\usepackage{amsmath}
\usepackage{mathtools}
\usepackage{amssymb}
\usepackage{subfigure}
\usepackage{booktabs}
\usepackage{caption}
\usepackage[capitalise,nameinlink]{cleveref}
\usepackage[normalem]{ulem}
\usepackage[makeroom]{cancel}

\definecolor{tablecolor}{rgb}{0.8,0.8,0.8}
\newcommand{\highlight}[1]{{\color{red} #1}}

\newcommand{\memory}{\mathcal{M}}
\newcommand{\setDist}{\mathcal{Q}}
\newcommand{\lik}{\text{lik}}
\newcommand{\cnst}{\text{const.}}

\newcommand{\losshat}{\hat{\ell}}
\newcommand{\tq}{\tilde{q}}

\newcommand{\qiso}{q^{\text{iso}}}
\newcommand{\qdiag}{q^{\text{diag}}}
\newcommand{\qfull}{q^{\text{full}}}
\newcommand{\losshatiso}{\hat{\ell}^{\text{iso}}}
\newcommand{\losshatdiag}{\hat{\ell}^{\text{diag}}}
\newcommand{\losshatfull}{\hat{\ell}^{\text{full}}}

\newcommand{\glmloss}{\mathcal{L}}

\newcommand{\kprior}{\mathcal{K}}
\newcommand{\hyp}{\delta}

\newcommand{\llm}{\text{pre}}
\newcommand{\ta}{\text{ta}}
\newcommand{\ba}{\text{ba}}
\newcommand{\ha}{\text{ha}}
\newcommand{\chk}{\text{chk}}

\newcommand{\jnt}{\text{jnt}}
\newcommand{\dualparamhat}{\hat{\boldsymbol{\lambda}}}
\newcommand{\site}{s}

\newcommand{\remove}{\backslash}

\newcommand{\diag}{\text{diag}}
\newcommand{\first}{{\text{first}}}
\newcommand{\second}{{\text{second}}}

\newcommand{\new}{\text{new}}
\newcommand{\old}{\text{old}}

\newcommand{\natgrad}{\widetilde{\nabla}}
\newcommand{\grad}{\nabla}

\newcommand{\entropy}{\mathcal{H}}

\newcommand{\loss}{\ell}
\newcommand{\vparam}{\boldsymbol{\theta}}
\newcommand{\param}{\theta}

\newcommand{\ty}{\tilde{y}}

\newcommand{\vnatparam}{\vlambda}

\newcommand{\vmeanparam}{\vmu}

\newcommand{\dkls}[3]{\mathbb{D}_{\text{KL}}^{#1}[#2 \, \|\, #3]}

\newcommand\cut[1]{}

\newcommand{\tm}{\widetilde{m}}

\newcommand{\squishlist}{
   \begin{list}{$\bullet$}
    { \setlength{\itemsep}{0pt}      \setlength{\parsep}{3pt}
      \setlength{\topsep}{3pt}       \setlength{\partopsep}{0pt}
      \setlength{\leftmargin}{1.5em} \setlength{\labelwidth}{1em}
      \setlength{\labelsep}{0.5em} } }

\newcommand{\squishlisttwo}{
   \begin{list}{$\bullet$}
    { \setlength{\itemsep}{0pt}    \setlength{\parsep}{0pt}
      \setlength{\topsep}{0pt}     \setlength{\partopsep}{0pt}
      \setlength{\leftmargin}{2em} \setlength{\labelwidth}{1.5em}
      \setlength{\labelsep}{0.5em} } }

\newcommand{\squishend}{
    \end{list}  }

\newtheorem{thm}{Theorem}{}
{}
{}

\newcommand{\half}{\mbox{$\frac{1}{2}$}}

\newcommand{\rnd}[1]{\left(#1\right)}
\newcommand{\sqr}[1]{\left[#1\right]}

\newcommand{\myang}[1]{\langle#1\rangle}
\newcommand{\myexpect}{\mathbb{E}}

\newcommand{\gauss}{\mbox{${\cal N}$}}

\newcommand{\myvec}[1]{\mbox{$\mathbf{#1}$}}
\newcommand{\myvecsym}[1]{\mbox{$\boldsymbol{#1}$}}

\newcommand{\vepsilon}{\mbox{$\myvecsym{\epsilon}$}}

\newcommand{\vmu}{\boldsymbol{\mu}}

\newcommand{\vlambda}{\boldsymbol{\lambda}}

\newcommand{\vSigma}{\mathbf{\Sigma}}

\newcommand{\va}{\mbox{$\myvec{a}$}}
\newcommand{\vb}{\mbox{$\myvec{b}$}}

\newcommand{\vh}{\mbox{$\myvec{h}$}}

\newcommand{\vm}{\mathbf{m}}

\newcommand{\vs}{\mbox{$\myvec{s}$}}

\newcommand{\vu}{\mbox{$\myvec{u}$}}
\newcommand{\vv}{\mbox{$\myvec{v}$}}

\newcommand{\vx}{\mbox{$\myvec{x}$}}

\newcommand{\vH}{\mathbf{H}}
\newcommand{\vI}{\mbox{$\myvec{I}$}}

\newcommand{\vS}{\mbox{$\myvec{S}$}}
\newcommand{\vT}{\mathbf{T}}

\newcommand{\trace}{\mbox{Tr}}

\newcommand{\calD}{\mbox{${\cal D}$}}

\newcommand{\data}{\calD}

\newcommand{\sigmoid}{\mbox{$\sigma$}}

\DeclareMathOperator*{\argmin}{arg\,min}
\crefname{section}{Sec.}{Sections}
\crefname{appendix}{App.}{Appendices}
\crefname{algorithm}{Alg.}{Algorithms}
\crefname{equation}{Eq.}{Eqs.}
\crefname{figure}{Fig.}{Figures}
\creflabelformat{equation}{#2\textup{#1}#3} 

\begin{document}

\title{Knowledge Adaptation as Posterior Correction}

\author{\name Mohammad Emtiyaz Khan \email emtiyaz.khan@riken.jp \\
       \addr RIKEN Center for Advanced Intelligence Project\\
       1-4-1 Nihonbashi, Chuo-ku, Tokyo 103-0027, Japan}

\editor{TBD}

\maketitle

\begin{abstract}
   Adaptation is the holy grail of intelligence, but even the best AI models lack the adaptability of toddlers.
   In spite of great progress, little is known about the mechanisms by which machines can learn to adapt as fast as humans and animals.
   Here, we cast adaptation as `correction' of old posteriors and show that a wide-variety of existing adaptation methods follow this very principle, including those used for continual learning, federated learning, unlearning, and model merging. 
   In all these settings, more accurate posteriors often lead to smaller corrections and can enable faster adaptation. Posterior correction is derived by using the dual representation of the Bayesian Learning Rule of \citet{khanRue23}, where the interference between the old representation and new information is quantified by using the natural-gradient mismatch.~We present many examples demonstrating how machines can learn to adapt quickly by using posterior correction.
   \end{abstract}


\section{Introduction}


Quick adaptation is essential for survival in nature so much so that it can be equated to intelligence~\citep{sternberg2019theory}. To this day, there is no Artificial Intelligence (AI) system whose adaptive abilities are considered anywhere close to those of humans and animals. Large Language Models (LLMs) like GPT and Gemini seem to do many tasks better than experts, but lack adaptability of toddlers. The only sense in which they adapt, through prompting, is temporary and
arguably superficial. We cannot teach new things to such models by interacting with them. 
There are no easy ways to quickly fix their mistakes and no magical tool to permanently add, remove, or modify their knowledge. The most trusted way is to simply retrain the whole system which is too costly to fix problematic behaviors reported on a daily basis.

\begin{figure}
   \center
   \includegraphics[width=6in]{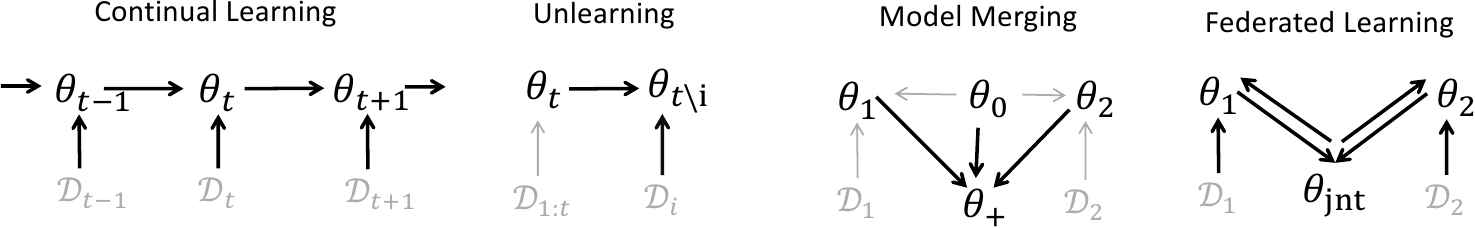}
   \caption{Four popular scenarios for adaptation of model parameters $\vparam$. Black arrows indicate the flow of adapted knowledge, while the gray arrows indicate pre-trained knowledge. Continual learning adapts $\vparam_t$ to $\vparam_{t+1}$ to include new data $\data_{t+1}$. Unlearning aims to estimate the model after removing a specific data set $\data_i$. Model merging attempts to improve a pre-trained base model $\vparam_0$ by merging back the fine-tuned models $\vparam_1$ and $\vparam_2$. Finally, federated learning aims to obtain a joint model $\vparam_\jnt$ by using locally trained models.}
   \label{fig:adaptation}
\end{figure}

A lot has been done recently to instill adaptivity in large AI models but these efforts are fragmented with no apparent similarities among the prominent methodologies (\cref{fig:adaptation}).~For example, continual-learning methods adapt to streaming data over time \citep{kirkpatrick2017overcoming}, while federated-learning methods assimilate knowledge spread across different locations \citep{mcmahan2016fedavg}. While both deal with distributed information, they use fundamentally different methodologies.
As an example, replay of old data
is common in continual learning \citep{rebuffi2017icarl,pan2020continual,buzzega2020dark} but discouraged in federated learning due to privacy concerns.
Such differences can make it harder to connect these two fields at a fundamental level.
The same is true of other approaches, such as, model merging \citep{wortsman22modelsoups, wortsman2021robust, IlharcoRWSHF23, DBLP:conf/iclr/MitchellLBFM22}, knowledge distillation
\citep{hinton2015distilling}, fine-tuning \citep{DBLP:conf/iclr/HuSWALWWC22}, etc. All such approaches attempt to adapt and reuse old knowledge quickly and avoid retraining as much as possible, but they remain disconnected from each other. 

Our goal in this paper is to unify these efforts and unravel a common adaptation mechanism behind them. We propose to formulate adaptation of old knowledge as `correction' of old posterior approximations. For example, in continual learning, inclusion of new data can be expressed as matching the predictions of the new and old posteriors. We give several such examples where knowledge adaptation, when formulated as posterior correction, amounts to matching the predictions, gradients, and/or
Hessians. 
Through examples in problems such as continual learning, unlearning, model merging, and federated learning, we argue that better posterior approximations reduce the correction needed to adapt and can facilitate faster adaptation. The proposal can also be useful for traditional approaches such as domain adaptation \citep{ben2006analysis} and semi-supervised learning \citep{olivier2006semi}, although we focus on recent efforts to instill adaptivity in large AI models.
Ultimately, posterior correction provides a new way to form and inject prior knowledge for faster adaptation, even in cases where there are no Bayesian models available or exact computation of the posterior is infeasible.

\section{Posterior Correction}

To derive posterior correction, we will use a variational-Bayesian framework that applies to a wide variety of learning problems, including Bayesian and non-Bayesian approaches. The goal of variational learning (VL) is to find good approximations of the posterior distribution \citep{khanRue23}, but it can also be seen as a generalization of the well-known Empirical Risk Minimization (ERM) problem. For instance, the ERM problem of optimizing over model parameters $\vparam \in \Theta$ can also be 
reformulated as a VL problem where we optimize over the distributions $q(\vparam) \in \setDist$. That is, we can write 
\begin{equation}
   \vparam_t = \argmin_{\vparam \in \Theta} \,\, \sum_{i=0}^t \loss_i(\vparam)  
   \qquad \text{as}\qquad
   q_t = \argmin_{q \in \setDist } \,\, \sum_{i=1}^t \myexpect_q[ \loss_i] + \dkls{}{q}{p_0} .
   \label{eq:ERMasVL}
\end{equation}
The ERM objective on the left uses loss functions $\loss_i$ defined over data sets $\data_i$ indexed by $i=1,2,\ldots, t$, and we denote the regularizer by $\loss_0$ and it can either be added explicitly, or obtained implicitly through a learning algorithm. In contrast to ERM, the VL problem uses the expected loss $\myexpect_{q}[\loss_i]$ under $q(\vparam)$ and a prior $p_0 \propto \exp(-\loss_0)$ in the Kullback-Leibler (KL) divergence term. Despite these differences, as shown in \cref{sec:adapt_main}, the ERM solution $\vparam_t$ can be
recovered as a special case of a VL problem in which $\setDist$ is the set of Gaussian distributions. In this case, the mean $\vm_t$ of the Gaussian $q_t$ is equal to $\vparam_t$ when we use the delta method to approximate the expectations $\myexpect_{q_t}[\loss_i] \approx \loss_i(\vm_t)$ in the VL objective.

Posterior correction uses the VL framework to cast knowledge adaptation as correction of old posteriors. We will now demonstrate the main idea for the problem of continual learning where the goal is to adapt to a new task $t+1$ an old model trained on $t$ previous tasks. In VL terms, this goal translates to the adaptation of the posterior $q_t$ shown in \cref{eq:ERMasVL} to obtain the posterior $q_{t+1}$, which additionally includes the loss on the new task (denoted by
$\loss_{t+1}$).
The optimal update is to use the Bayes' rule, that is, we compute the new posterior $p_{t+1}$ as follows: $p_{t+1} \propto q_t \exp\rnd{-\loss_{t+1}}$ where $q_t$ is the prior and $\loss_{t+1}$ defines the likelihood. However, this is challenging unless the loss and prior form a conjugate pair, like Gaussians \citep[p. 117]{bishop2006pattern}. A tractable alternative is Variational Continual Learning \citep{cuong2018}, which uses the following recursive variational form setting the prior to $q_t$,
\begin{equation}
   \hat{q}_{t+1} = \argmin_{q \in \setDist } \,\, \myexpect_q[ \loss_{t+1} ] + \dkls{}{q}{q_t}.
   \label{eq:vcl}
\end{equation}
Unfortunately, this does not exactly recover the true $q_{t+1}$. Soon we will see that exact recovery is possible if we include a \emph{correction} term in the above update.

\begin{figure}[!t]
   \center
   \includegraphics[width=6in]{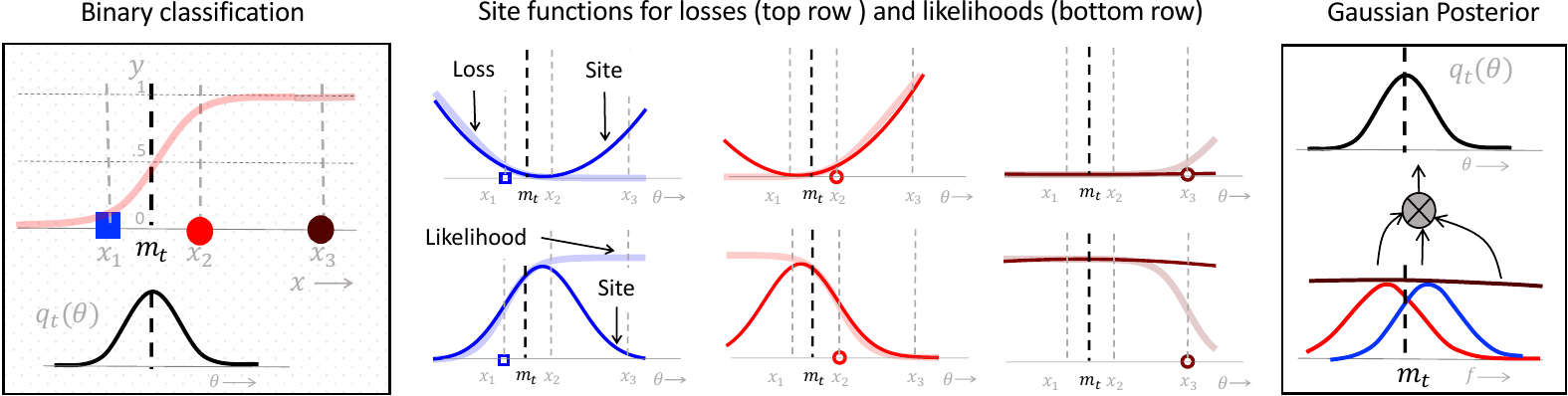}
   \caption{We illustrate the dual form given in \cref{eq:qdualform} on a 1-D binary classification example (circle vs square in the leftmost panel). The classifier is simply a threshold $\theta$, fit via a logistic likelihood (pink sigmoid). We use a Gaussian $q_t(\theta)$ with mean $m_t$ shown with vertical dashed line. The logistic losses for the three examples are shown in the top row of middle panel (transparent lines) along with their quadratic sites (solid lines). The bottom row
      displays likelihoods. The rightmost panel applies \cref{eq:qdualform} to form $q_t$. The example farthest from the classifier has the smallest curvature and least contribution to $q_t$. The example is discussed in detail in \cref{app:oneDex} along with a general case of logistic regression in \cref{app:logreg}.}
   \label{fig:dual_form}
\end{figure}

The correction term can be derived by using a \emph{dual} representation of $q_t$, originally proposed by \citet{khan2018fast1}. Essentially, we can write $q_t$ as the product of local factors defined using \emph{site} functions for each $\loss_i$. The sites are essentially surrogates of the loss which often take a simple form; see an  illustration in \cref{fig:dual_form}. More precisely, let $\setDist$ be the set of minimal exponential-family distributions $q(\vparam) =
\exp(\myang{\vT(\vparam), \vnatparam})/\mathcal{Z}$ with sufficient statistics $\vT(\vparam)$, natural
parameter $\vlambda$, and partition function $\mathcal{Z}$; the inner product is denoted by $\myang{\cdot,\cdot}$. 
Then, a solution $q_t$ of \cref{eq:ERMasVL} can be written as a product of local factors as follows, 
\begin{equation}
   q_{t}(\vparam) = \frac{1}{\mathcal{Z}_t} \prod_{i=0}^t \exp\rnd{-\losshat_{i|t}(\vparam)}, 
   \,\,\text{ where } \losshat_{i|t}(\vparam) = \myang{\vT(\vparam) , \natgrad \myexpect_{q_t}[\loss_i]}.
   \label{eq:qdualform}
\end{equation}
By $\losshat_{i|t}$ we denote the site function for loss $\loss_i$ evaluated at $q_t$. The expression of the site function shows that they are obtained by using \emph{natural gradients}, denoted by $\smash{\natgrad \myexpect_{q_t}[\loss_i]}$, taken with respect to the natural parameter $\vnatparam_t$ of $q_t$. A definition of natural gradients and detailed derivation of the above expression are both provided in \cref{sec:dualBLR}, along with a generalization to the iterates of the Bayesian
learning rule (BLR) \citep{khanRue23} which is later used to boost learning trajectories.

\begin{figure}[!t]
   \centering
   \begin{subfigure}
      \centering
      \includegraphics[height=2.0in]{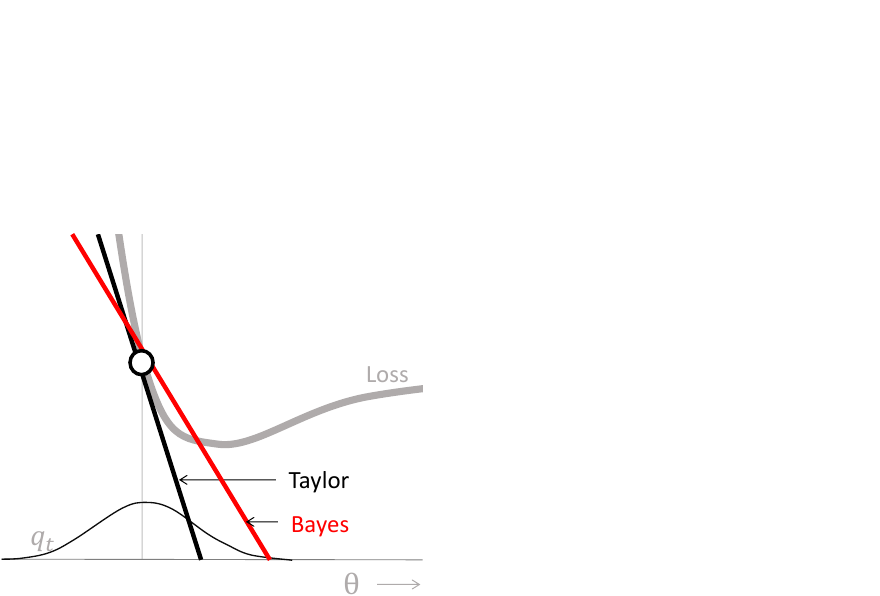}
   \end{subfigure}
   \hspace{1cm}
   \begin{subfigure}
      \centering
      \includegraphics[height=2.0in]{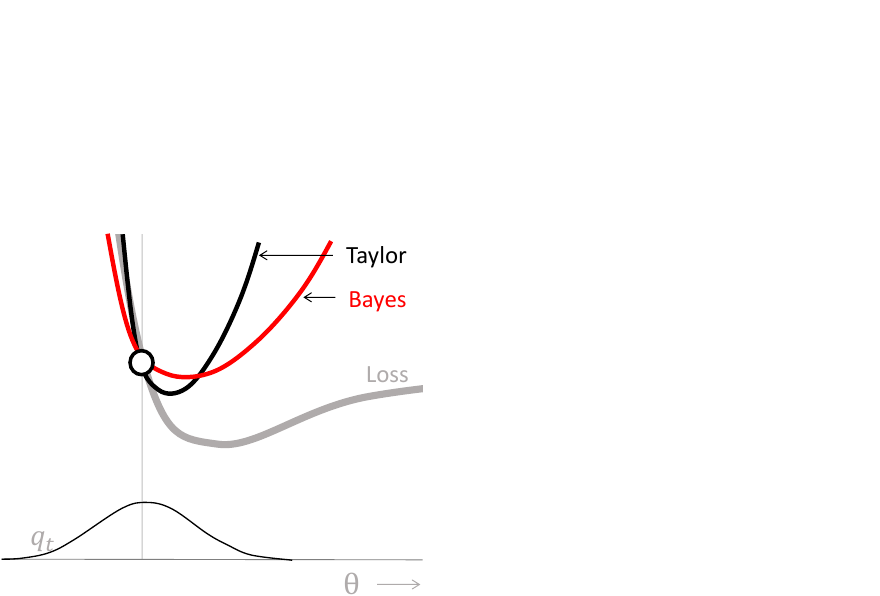}
   \end{subfigure}
   \caption{The left panel compares the site \smash{$\losshatiso_{i|t}$} to the $\first$-order Taylor expansion of $\loss_i$ at $\vm_t$. The right panel compares \smash{$\losshatfull_{i|t}$} to the $\second$-order Taylor. The sites use expectations of the gradients and Hessians over $q_t$ and
   capture more global information around $\vm_t$.}
   \label{fig:sites}
\end{figure}

\cref{tab:site} lists some examples of the sites which will be used later in the paper. We note that, for Gaussian posteriors, the sites resemble Taylor's approximations of the loss. The main difference is that the sites use derivatives evaluated and averaged over samples from $q_t$; see \cref{fig:sites}. The site functions contain more global information than Taylor surrogates and they apply more generally, for instance, to  discontinuous losses over discrete variables.

\begin{table}[!t]
   \begin{center}
      \caption{Site functions. We denote $\bar\vH_{i|t} = \myexpect_{q_t}[\nabla^2 \loss_i]$ and its diagonal by the vector $\bar\vh_i$. Element-wise multiplication and divisions are denoted by $\va \cdot \vb$ and $\va/\vb$ respectively.}
      \label{tab:site}
      \begin{tabular}{lll}
         \toprule
         Form of $q_t$ & Expression for the site $\losshat_{i|t}$ \\
         \bottomrule
         Gaussian $\qiso_t = \gauss(\vparam|\vm_t, \vI)$ & $\losshatiso_{i|t} =  \vparam^\top \myexpect_{q_t}[\nabla \loss_i]$ \\
         Gaussian $\qdiag_t =  \gauss(\vparam|\vm_t, \diag(\frac{1}{\vs_t}))$ &  $\losshatdiag_{i|t} = \vparam^\top \myexpect_{q_t}[\nabla \loss_i] + \half (\vparam-\vm_t)^\top \sqr{ \bar\vh_i \cdot (\vparam-\vm_t) } $\\
         Gaussian $\qfull_t = \gauss(\vparam|\vm_t, \vS_t^{-1})$ & $\losshatfull_{i|t} = \vparam^\top \myexpect_{q_t}[\nabla \loss_i] + \half (\vparam-\vm_t)^\top \bar\vH_{i|t} (\vparam-\vm_t) $\\
         \text{Bernoulli} $P(\param = 1) = p_1$ & $\losshat^{\text{bern}}_{i|t} = \param\,\, \nabla_{p_1} \myexpect_{q_t}[\loss_i]$ \\ 
         \bottomrule
      \end{tabular}
   \end{center}
\end{table}

   We use \cref{eq:qdualform} to correct the update in \cref{eq:vcl} to recover $q_{t+1}$ exactly, as shown in the equation below. In the first line, we begin with the definition of $q_{t+1}$ from \cref{eq:ERMasVL}. In the second line, we expand the KL term and divide and multiply by $q_t$. In the third line, we rewrite the $q_t$ that appears in the numerator in its dual form, and finally we rearrange terms to get the correction term in the last line,
\begin{equation}
   \begin{split}
      q_{t+1} &= \argmin_{q \in \setDist } \,\, \sum_{i=1}^{t+1} \myexpect_q[ \loss_i] + \dkls{}{q}{p_0}  \\
      &= \argmin_{q \in \setDist } \,\, \myexpect_q[ \loss_{t+1}] + \sum_{i=1}^t \myexpect_q[ \loss_i] + \myexpect_q \sqr{ \log \rnd{\frac{q}{p_0} \times \frac{q_t}{q_t} } }   \\
      &= \argmin_{q \in \setDist } \,\, \myexpect_q[ \loss_{t+1}] + \sum_{i=1}^t \myexpect_q[ \loss_i] + \myexpect_q \sqr{ \log \rnd{\frac{q}{\frac{1}{\mathcal{Z}_0}e^{-\loss_0}} \times \frac{\frac{1}{\mathcal{Z}_t}\prod_{i=0}^t e^{-\losshat_{i|t}}}{q_t} } }   \\
      &= \argmin_{q \in \setDist } \,\, \myexpect_q[ \loss_{t+1}] + \dkls{}{q}{q_t} + \sum_{i=0}^t \underbrace{ \myexpect_q[ \loss_i - \losshat_{i|t}] }_{\text{Correction}} 
   \end{split}
   \label{eq:postcorr}
\end{equation}
The third term in the last line corrects the update of \cref{eq:vcl}, ensuring the exact recovery of $q_{t+1}$ (instead of the approximate $\hat{q}_{t+1}$). Corrections are required for all the past $\loss_i$'s as well as for the regularizer $\loss_0$. Essentially, past $\loss_i$'s are represented in $q_t$ through sites \smash{$\losshat_{i|t}$} and, when a new $\loss_{t+1}$ is added, the site $\losshat_{i|t}$ needs to change to \smash{$\losshat_{i|t+1}$}.
   The correction term gives a precise mathematical characterization of this \emph{interference} between the old representation and new information, which is an age-old problem \citep{sutton1986two} without a precise agreed-upon mathematical definition. Other views of posterior correction are discussed in \cref{sec:postcorr_details} where the correction is expressed as a Bayesian filter, a new definition of information gain is given, and interference is interpreted as
   the natural-gradient mismatch computed over the past data. In addition, we discuss a method to boost training trajectories.

\section{Knowledge Adaptation as Posterior Correction}
   \label{sec:adapt_main}

The main result of this paper is to show that many existing adaptation methods tailored to handle specific adaptation cases, can all be seen as specific instances of posterior correction. This key message is demonstrated by covering multiple adaptation scenarios, such as, continual learning, unlearning, model merging, and federated learning. The results are derived by relying on the following
three types of Gaussian families defined in \cref{tab:site}: 
\begin{enumerate}
   \item The `isotropic Gaussian' family is denoted by candidates $\qiso = \gauss(\vparam|\vm, \vI)$ where the mean $\vm$ is learned but the covariance is fixed to the identity matrix $\vI$;
   \item The `diagonal Gaussian' family denoted by candidates $\qdiag =  \gauss(\vparam|\vm, \diag(\vs)^{-1})$ where the mean $\vm$ and the diagonal vector $\vs$ of the precision matrix are both learned;
   \item The `full Gaussian' family denoted by $\qfull = \gauss(\vparam|\vm,
\vS^{-1})$ where $\vm$ and $\vS$ are learned.
\end{enumerate}
We will see that increasing the size of the family from $\qiso$ to $\qdiag$ and then to $\qfull$ often leads to reductions in corrections. Smaller corrections imply less interference and yield better adaptation methods. This message is repeated throughout the paper through many examples, all aimed at establishing posterior-correction as a unifying principle for knowledge adaptation.
We do not derive a precise relation between the size of corrections and speed of adaptation. That would require in-depth studies for each case separately which is out of the scope for this work. Our hope is that the future work will benefit from the insights of this paper to perform such studies for different types of adaptation tasks.

   Throughout, we will reformulate an ERM problem as a VL problem. For instance, for a $\qiso$ family, VL in \cref{eq:ERMasVL} reduces to an ERM over the mean $\vm$ by using the $\first$-order delta method \citep[App. C]{khanRue23},
\begin{equation}
   \myexpect_{q}[\loss_i] \approx 
   \loss_i(\vm) 
   \quad\implies\quad
   \vm_t = \argmin_{\vm\in\Theta} \,\, \sum_{i=0}^t \loss_i(\vm) + \cnst
   \label{eq:deltamethod}
\end{equation}
This is because the entropy of $\qiso$ is constant, therefore the KL term in VL only contributes the  $\loss_0$ term. This reformulation is suitable for ERMs that aim for the $\first$-order stationarity condition, for instance, by using stochastic gradient descent. For algorithms that aim for $\second$-order optimality, we can use a $\second$-order delta method over a $\qfull$ family and assume that the covariance is set to the Hessian at the ERM solution $\vm_t$ which corresponds to Laplace's
method \citep[App. C.1]{khanRue23}. We start with the first adaptation task of continual learning in the next section.

\subsection{Continual Learning as Posterior Correction}

In continual learning, the goal is to quickly update $\vparam_t$ to $\vparam_{t+1}$ when a new loss $\loss_{t+1}$ is available. A popular strategy for this is to use regularization.
Three kinds of regularization methods are commonly employed: (i) parameter-space regularization, (ii) function or prediction-space regularization, (iii) experience or memory replay.
We will show that such regularizers naturally arise as special cases of the regularizer used in \cref{eq:postcorr} for posterior correction. The regularizer is shown below and denoted by
\begin{equation}
   \kprior_t(q) = \dkls{}{q}{q_t} + \sum_{i=0}^t \myexpect_q[ \loss_i - \losshat_{i|t}].
   \label{eq:varkprior}
\end{equation}
The KL regularizer operates in the parameter space and the correction term adds further regularization in the data-space, containing as special cases the function and prediction regularization, as well as the experience and memory replay methods. Overall we will show that posterior correction naturally combines all these regularizers into a single regularizer. 

\subsubsection{Parameter-space regularization as Posterior Correction}

We start with the parameter-space regularization which is implemented in $\kprior_t(q)$ through the KL regularizer. Dropping the correction term, simply reduces to \cref{eq:vcl} which is known as Variational Continual Learning \citep{cuong2018}. Different choices of distribution give rise to different regularizers in the parameter space.
For instance, for Gaussian families, the KL term reduces to a quadratic regularizer, shown below for the $\qiso$ and $\qfull$ families,
\begin{align}
   \dkls{}{\qiso}{\qiso_t} &=  \half (\vm-\vm_t)^\top(\vm-\vm_t) + \cnst\label{eq:kl} \\
   \dkls{}{\qfull}{\qfull_t} &=  \half (\vm-\vm_t)^\top\vS_t(\vm-\vm_t) + \half \trace\sqr{\vS^{-1}\vS_t} - \half \log\left|\vS^{-1}\vS_t\right| + \cnst \nonumber
\end{align}
The curvature of the regularizer is controlled through the posterior covariance, for instance, using the family $\gauss(\vparam|\vm, \vI/\delta)$ with a scalar precision $\delta<0$ will lead to the weighting of the quadratic term by $\delta$. Similarly, for the $\qfull$ family, the curvature can be set to the Hessian if we invoke the delta method thereby recovering the so-called Laplace approximation, as detailed in \citet[App. C]{khanRue23}. A similar choice with the $\qdiag$ family
recovers the popular Elastic-Weight Consolidation \citep{kirkpatrick2017overcoming}. In general, employing an arbitrary exponential-family, the KL term enables generic parameter-space regularizations as in online learning \citep{hoeven2018many}.

\subsubsection{Function-space regularization as Posterior Correction}

\begin{figure}[t!]
   \center
   \includegraphics[width=2.3in]{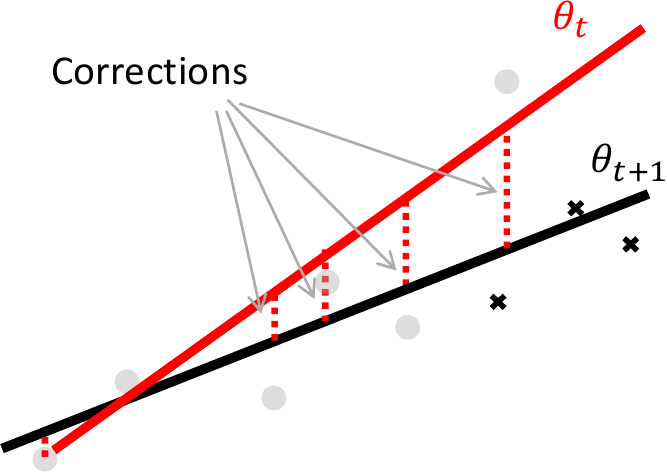}
   \hspace{1cm}
   \includegraphics[width=2.3in]{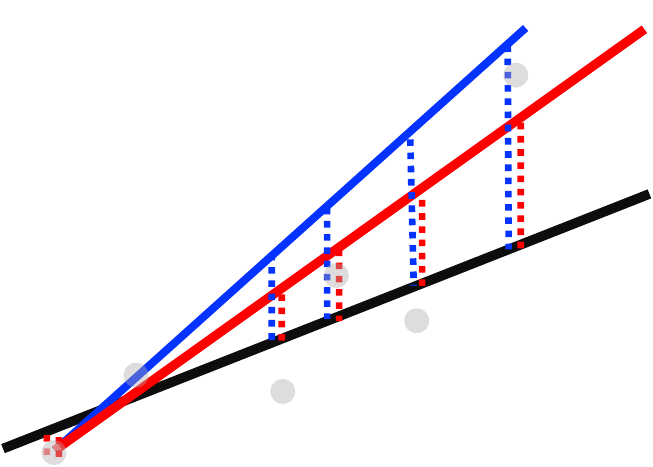}
   \caption{Visualization of corrections as prediction mismatches for linear regression with $\qiso$ family. The old model is trained on the gray `o' and the new model additionally includes the black `$\times$' too. Corrections are mismatches over old examples (dashed, vertical red lines). The right figure shows an additional slightly-worse model (in blue) with larger mismatches. The red model has smaller correction because it is closer to the black model which also supports
the intuition that smaller corrections imply faster adaptation.}
   \label{fig:linreg_corr}
\end{figure}

The correction term adds further regularization on top of the KL term, which encompasses function- and prediction- space regularizations. A simple example illustrating this is the squared loss with a linear predictor $\hat{y}_i(\vparam) = \vx_i^\top \vparam$ for a given input-output pair $(\vx_i, y_i)$,
\[
   \loss_i(\vparam) = \glmloss\sqr{ y_i, \, \hat{y}_i(\vparam) } = \half \| y_i - \hat{y}_i(\vparam)\|^2.
\] 
For this loss, the correction term simplifies as `prediction matching' among the new and old model when we use a $\qiso$ family. This is shown below:
\begin{equation}
   \begin{split}
      \loss_i(\vparam) - \losshatiso_{i|t}(\vparam) &= \loss_i(\vparam) - \vparam^\top \myexpect_{q_t}[\nabla \loss_i] \\
      &= \half \|y_i - \vx_i^\top \vparam\|^2 - \vparam^\top \vx_i (\vx_i^\top \vm_t - y_i) \\
      &= \half \| \vx_i^\top\vm_t -  \vx_i^\top\vparam \|^2 + \cnst 
   \end{split}
   \label{eq:corr_linreg}
\end{equation}
This is equivalent to function-space regularization \citep{benjamin2018measuring}. It also corresponds to Knowledge Distillation \citep{hinton2015distilling} under large temperature and centered logits. A visualization is shown in \cref{fig:linreg_corr} to highlight the intuition that a model with smaller corrections is also closer to the final goal, and therefore can facilitate faster adaptation.


\subsubsection{Prediction Matching as Posterior Correction of Neural Networks}

The result extends to more general cases where corrections take a form similar to Knowledge Distillation, but also account for errors arising due to the model's non-linearity. To see this, we consider a general nonlinear model such as a neural network, denoted by $f_i(\vparam)$ and trained by penalizing its predictions $\hat y_i$ through a Bregman loss $\glmloss(y_i, \, \hat yi)$. For example, we can use the cross-entropy loss where $\hat y_i$ are obtained by passing
$f_i(\vparam)$
through the softmax function. We show in \cref{sec:cl_corr_term} that, for such cases, the correction can be written as prediction-matching
with an additional error term arising due to non-linearity,
\begin{equation}
   \loss_i(\vparam) - \losshatiso_{i|t}(\vparam) \,\, =  \,\, \underbrace{ \glmloss\sqr{\hat{y}_{i|t}, \, \hat{y}_i(\vparam) } }_{\text{Pred. matching}} + r_{i|t} \underbrace{ \| f_i(\vparam) - \hat{f}^{\text{lin}}_i(\vparam) \|^2 }_{\text{Linearization error}} + \epsilon_{i|t} .
   \label{eq:pred_match_general}
\end{equation}
The first term takes a form similar to distillation where we denote by $\hat y_{i|t} = \myexpect_{q_t}[ \hat{y}_i(\vparam)]$ the average posterior predictions of $q_t$. The sampling can be seen as playing a role similar to the temperature. 
In the second term, we denote the residual by $r_{i|t} = \hat y_{i|t} - y_i$ and the $\first$-order linearization of $f_i(\vparam)$ at $\vm_t$ by \smash{$\hat f^{\text{lin}}_i(\vparam) = f_i(\vm_t) + (\vparam - \vm_t)^\top \nabla f_i(\vm_t)$}.
The linearization error disappears when the model is linear, leading to a purely mismatch-based correction as in
\cref{eq:corr_linreg}. The $\epsilon_{i|t}$ is an error term that arise due to the use of the delta method to approximate the expectation over $q_t$ at its mean $\vm_t$. The error is expected to be small and can be safely ignored.

Overall, the expression shows that the past mistakes are propagated due to non-linearity, and therefore cause no issues at all for linear cases where prediction matching would be sufficient for accurate continual learning. For non-linear models, past mistakes can hurt through examples where the linearization error in $f_i(\vparam)$ is large. For such examples, the correction term suggests to avoid abrupt changes in the feature. 

\subsubsection{How to reduce corrections and enable accurate continual learning?}
A straightforward way to reduce the correction is by using a more accurate posterior. We show this in \cref{sec:cl_corr_term}. By using the $\qfull$ family, the $\hat
f_i^{\text{lin}}(\vparam)$ is replaced by a quadratic surrogate $\hat f_i^{\text{quad}}(\vparam) = \hat f_i^{\text{lin}}(\vparam) + \half (\vparam-\vm_t)^\top \nabla^2 f_i(\vm_t) (\vparam-\vm_t)$. A quadratic surrogate yields a smaller linearization error, thereby reducing the harm caused by past mistakes. In fact, for a
squared loss, the correction term completely vanishes, giving rise to the classical Bayes' update for linear regression, where exact continual learning is possible just with the KL term. The example demonstrates that the need for correction is reduced with more flexible posteriors, which can speed up the adaptation process.

Flexible posteriors however can be costly. For example, computing full covariances is infeasible for larger neural networks. For such cases, \cref{eq:pred_match_general} suggests a cheaper alternative, which is to simply use replay of past examples to improve continual learning.
The split of errors shown in \cref{eq:pred_match_general} motivates the use of mixing knowledge distillation with experience replay, as done in Dark Experience Replay \citep{buzzega2020dark}.
Another approach is derived in \citet{daxberger2023improving} by using the K-prior framework of \citet{khan2021knowledge}, where they combine the three types of regularization methods. We show in \cref{sec:kprior_details} that the K-prior, which also involves prediction matching, is a special case of posterior correction with the $\qiso$ family. Posterior correction generalizes the K-prior by allowing the use of an arbitrary posterior distribution. This extends the prediction matching to
other kinds of matching, such as, gradient and Hessian matching. It also provides a simpler way to mix and match various regularization methods as discussed in \cref{sec:kprior_details}. 

In \cref{sec:kprior_details}, we also briefly discuss connections to other approaches covered under the K-prior framework, such as incremental Support Vector Machines \citep{cauwenberghs2001incremental, liang2009incremental}, Similarity Control \citep{vapnik2015learning}, and memory-based continual learning \citep{lopez2017gradient}.
Finally, we show an application to sequential Bayesian inference by deriving a method by \citet{bui2017streaming} for sequential Sparse Variational Gaussian Process as a special instance of posterior correction.

\subsection{Unlearning as Posterior Correction}

Unlearning aims to recover the model parameters obtained after removing a given data set $\data_i$. A specific case is the influence estimation which focuses on estimating the change in the model when a specific data point is removed, but the same idea can also be applied to group of examples. In this sense, the goal of unlearning and influence are similar. More concretely, given a model $\vparam_t$ trained on $t$ data examples, we may want to estimate the new model
$\vparam_{t \remove j}$ trained on all the data except the $j$'th example. Unlearning aims to arrive at $\vparam_{t \remove j}$ without retraining from scratch. Influence estimates provide a simple expression for this, often obtained using a Taylor approximation evaluated at $\vparam_t$. 
We will now derive these estimates by using posterior correction. Specifically, we will show two results to derive TracIn \citep{pruthi2020estimating} and Influence Function (IF) \citep{jaeckel1972infinitesimal,cook1977detection, koh2017understanding}. 

We start by rewriting the influence calculation as posterior correction. Using a variational reformulation, we assume that $q_t$ is given and our goal is to estimate the posterior $q_{t \remove j}$ trained on all data examples except the $j$'th one. This is shown below in the first line. The following lines are obtained similarly to \cref{eq:postcorr} to get the correction,
\begin{equation}
   \begin{split}
      q_{t\remove j} &= \argmin_{q \in \setDist } \,\, \myexpect_q[-\loss_j] + \sum_{i=1}^t \myexpect_q[\loss_i] + \dkls{}{q}{p_0} \\
            &= \argmin_{q \in \setDist } \,\, \myexpect_q[-\loss_j] + \dkls{}{q}{q_t} + \sum_{i=0}^t \myexpect_q[\loss_i - \losshat_{i|t}] \\
            &= \argmin_{q \in \setDist } \,\, \myexpect_q[-\losshat_{j|t}] + \dkls{}{q}{q_t} + \sum_{i=0:t, i\ne j} \myexpect_q[\loss_i - \losshat_{i|t}] \\
   \end{split}
   \label{eq:postcorr_influence}
\end{equation}
The second line characterizes the interference caused by the removal of $\ell_j$ and the third line rearranges the terms corresponding to the $j$'th example.

\subsubsection{TracIn as Posterior Correction}
The TracIn estimator uses the gradient $\nabla \loss_j(\vparam_t)$ as the influence estimator. This can be derived as a simplification of \cref{eq:postcorr_influence} where we make two changes. First, we restrict the set of distributions to the $\qiso$ family, denoted by $\setDist^{\text{iso}}$, with candidates $q = \gauss(\vparam|\vm, \vI)$. Second, we ignore the correction term.
With these approximations, TracIn is recovered from posterior correction as shown below, where in the first line we make the two changes, then in the second line we rewrite the objective as an optimization problem over $\vm$, and in the last line we show the solution:
\begin{equation}
   \begin{split}
      &q_{t\remove j} \approx \argmin_{q \in \setDist^{\text{iso}}}\,\,  \myexpect_q[-\losshat_{j|t}] + \dkls{}{q}{q_t} \\ 
      &\implies \vm_{t\remove j} \approx \argmin_{\vm}\,\,  \vm^\top \myexpect_{q_t}[-\nabla \loss_j] + \half \|\vm - \vm_t\|^2 \\
      &\implies \vm_{t\remove j} \approx \vm_t + \nabla \loss_j(\vm_t).
   \end{split}
   \label{eq:tracin_deriv}
\end{equation}
The last line is obtained by using the delta method to approximate $\myexpect_{q_t}[-\nabla \loss_j] \approx -\nabla \loss_j(\vm_t)$. Renaming $\vm_t$ to $\vparam_t$, we get the TracIn estimator $\vparam_{t\remove j} -\vparam_t \approx \nabla \loss_j(\vparam_t)$.
The derivation shows the approximations needed to derive TracIn from posterior correction. Next, we will show that by relaxing these approximations we can get a better influence estimator. Specifically, we will derive the influence function (IF) in a similar fashion but reducing the approximation errors introduced above. 

\subsubsection{Influence Function as Posterior Correction}
\label{sec:if_pc}
Unlike TracIn, the IF method uses a Newton step, but this can also be derived through a similar procedure by tightening the approximations made. Specifically, instead of completely ignoring the correction term, we approximate it by using the $\second$-order Taylor expansion
\[
   \loss_i(\vparam) \approx \loss_i(\vm_t) + (\vparam-\vm)^\top \nabla \loss_i(\vm_t) + \half (\vparam - \vm_t)^\top \nabla^2 \loss_i(\vm_t) (\vparam - \vm_t).
\]
Using this, the correction term is approximated by a quadratic. To simplify the derivation, we assume $\loss_0 = \half \delta \|\vparam\|^2$ for a scalar $\delta>0$, although the derivation holds more generally. For this case, we show in \cref{app:inf_corr_l0}, that the correction for the $\qiso$ family simplifies to
\begin{equation}
   \begin{split}
      \loss_0(\vparam) - \losshatiso_{0|t}(\vparam) &=  \loss_o(\vparam) - \vparam^\top \myexpect_{q_t}[\nabla \loss_0]  - \dkls{}{q}{q_t} + \cnst \\
   \end{split}
   \label{eq:inf_corr_l0}
\end{equation}
Using this, we can write the correction term as a quadratic,
\begin{equation}
   \begin{split}
      \sum_{i=0, i\ne j}^t \myexpect_q[\loss_i - \losshat_{i|t}] &= \sum_{i=0, i\ne j}^t \myexpect_q \sqr{ \loss_i(\vparam) - \vparam^\top \myexpect_{q_t}[\nabla \loss_i] } - \dkls{}{q}{q_t}+ \cnst\\
      &\approx \sum_{i=0, i\ne j}^t\myexpect_q \sqr{ \half (\vparam - \vm_t)^\top \nabla^2 \loss_i(\vm_t) (\vparam - \vm_t)} - \dkls{}{q}{q_t} + \cnst\\
      &= \half (\vm - \vm_t)^\top \vH_{t\remove j} (\vm - \vm_t) - \dkls{}{q}{q_t} + \cnst
   \end{split}
   \label{eq:corr_with_taylor}
\end{equation}
The second line is obtained by using the delta method $\myexpect_{q_t}[\nabla \loss_i] \approx \nabla \loss_i(\vm_t)$ which leads to the cancellation of the linear terms, leaving only the quadratic terms. The last line then follows by denoting the Hessian by \smash{$\vH_{t\remove j} = \sum_{i=0, i\ne j}^t \nabla^2 \loss_i(\vm_t)$}.

Using the above approximation, \cref{eq:postcorr_influence} simplifies to the influence function (IF),
\begin{equation}
   \begin{split}
      \vm_{t\remove j} &\approx \argmin_{\vm}\,\, \vm^\top \myexpect_{q_t}[-\nabla \loss_j] + \half (\vm-\vm_t)^\top \vH_{t\remove j} (\vm - \vm_t) \\
      &= \vm_t + \vH_{t\remove j}^{-1} \myexpect_{q_t} [\nabla \loss_j] \\
      &\approx \vm_t + \vH_{t\remove j}^{-1} \nabla \loss_j(\vm_t),
   \end{split} 
   \label{eq:if_deriv}
\end{equation}
where, similarly to \cref{eq:tracin_deriv}, we use the delta method to approximate the expectation over $q_t$. This is the IF for the removal of an example, for example, see \citet[Eq. 2]{nickl2023memory}. The derivation clearly shows the approximations used by TracIn and IF: the former ignores the corrections while the latter reduces it by using Taylor's approximation. IF is a better estimator, at least in theory, because it uses the Hessian as well. This supports our hypothesis that smaller corrections lead to
better adaptation methods.

An alternate way to derive the same result is to expand the size of the family from $\qiso$ to $\qfull$, which also leads to a reductions in corrections. As shown in \cref{app:if_qfull}, posterior correction in \cref{eq:postcorr_influence} yields the following update over the $\qfull$ family,
\begin{equation}
   \vm_{t\remove j} \approx \vm_t + \hat \vS_{t\remove j}^{-1} \myexpect_{q_t} \sqr{ \nabla \loss_j} 
   \text{ where } \hat \vS_{t\remove j} = \vS_t - \myexpect_{q_t} [\nabla^2 \loss_j].
   \label{eq:if_qfull}
\end{equation}
This reduces to \cref{eq:if_deriv} if we use the delta method to approximate $\myexpect_{q_t} \sqr{ \nabla \loss_j} \approx \nabla \loss_j(\vm_t)$ and \smash{$\myexpect_{q_t} \sqr{ \nabla^2 \loss_j} \approx \nabla^2 \loss_j(\vm_t)$}, and also use $\vS_t = \vH_t$. This shows that we do not have to explicitly approximate the correction (as we did before by using Taylor's method). We could obtain the same result by simply expanding the size of the posterior class. Both
methods lead to similar reductions of correction, and ultimately both lead to the same adaptation methods which is a better estimator than TracIn.

\subsubsection{Memory Perturbation as Posterior Correction}

The expression in \cref{eq:if_qfull} is in fact connected to a more general way of deriving influence estimators. A recent proposal by \citet{nickl2023memory} called the Memory-Perturbation Equation generalizes influence estimation by using the Bayesian Learning Rule (BLR). Their approach can also be seen as a special instance of posterior correction where a generic exponential family is used but the correction term is ignored. Let us denote the natural parameters of $q_t$ and $q_{t\remove j}$ by $\vlambda_t$ and $\vlambda_{t\remove j}$, respectively. Then, by using the definition of
$\losshat_{t|j}$ from \cref{eq:qdualform} and ignoring the correction term, we can rewrite \cref{eq:postcorr_influence} as follows:
\begin{equation}
   \vlambda_{t\remove j} \approx \argmin_{\vmu} \,\, -\myang{\vmu , \natgrad \myexpect_{q_t}[\loss_i]} + \dkls{}{q}{q_t}  
   \quad\implies \quad
   \vlambda_{t\remove j} \approx \vlambda_t + \natgrad \myexpect_{q_t}[\loss_i].
\end{equation}
The second expression is obtained by simply taking the derivative with respect to $\vmu$ and setting it to zero. The derivation simplifies by noting that the derivative $\nabla_{\vmu} \dkls{}{q_{t\remove j}}{q_t} = \vlambda_{t\remove j} - \vlambda_t$ \citep[Eq. 23]{khanRue23}. The last expression above is precisely the Memory-Perturbation Equation shown in \citet[Eq. 6]{nickl2023memory}. It is easy to check that when we use the $\qfull$ family the equation above reduces to \cref{eq:if_qfull}. The
memory-perturbation equation gives a general way to expand the class of the variational family to get better estimators, and as shown above it is a special instance of posterior correction where corrections are simply ignored. Overall, these connections suggests two ways to improve influence estimation: either we can expand the posterior-family class, or we could explicitly reduce the correction.

\subsection{Model Merging as Posterior Correction}

Model merging is a popular technique to improve capabilities of LLMs without retraining them from scratch. Given a pre-trained model with parameters $\vparam_\llm$ on loss $\loss_\llm$, we `fine-tune' models $\vparam_i$ on individual tasks $\loss_i$. For instance, given an LLM trained on English text, we may fine-tune it on other languages and merge them all to get a model that can also generate texts in other languages. The fine-tuning procedure is essentially
identical to the standard
stochastic training but is initialized at $\vparam_\llm$ and trained only on $\loss_i$ with early stopping to avoid overfitting. Given fine-tuned models $\vparam_i$ for all tasks $i$, the merging is done via a simple addition,
for instance, Task-Arithmetic (TA) \citep{IlharcoRWSHF23} uses the following merging (with weights $\alpha_i>0$):
\begin{equation}
   \vparam_\ta = \vparam_\llm + \sum_{i=1}^t \alpha_i (\vparam_i - \vparam_\llm).
   \label{eq:ta}
\end{equation}
This simple merging works extremely well for unknown reasons. In what follows, we will derive it as a special case of posterior correction where correction is ignored, and then show that an explicit reduction of the correction leads to a better merging method.

\subsubsection{Posterior Merging via Bayesian Arithmetic}
Similarly to other sections, we start by formulating model merging as a variational problem. We denoteby $q_\llm$ the pre-trained posterior on $\loss_\llm$ and by $q_i$ the posterior obtained by fine-tuning $q_\llm$ over the task $\loss_i$. For fine-tuning, we use the following variational objective as a proxy where the KL term is used to mimic the initialization used to warm start,
\begin{equation}
   q_i = \argmin_{q\in\setDist^{\text{iso}}} \,\, \myexpect_q[\loss_i] + \dkls{}{q}{q_\llm}.
   \label{eq:qi}
\end{equation}
Given $q_\llm$ and all $q_i$, our goal is to obtain the model trained from scratch on all the data jointly. A simple rule to merge the posteriors $q_i$ is the following:  
\begin{equation}
   q_\ba =  q_\llm\prod_{i=1}^t \rnd{ \frac{q_i}{q_\llm}}^{\alpha_i},
   \label{eq:ba}
\end{equation}
which has a similar form to \cref{eq:ta} if we replace addition and subtraction by multiplication and division. Due to this analogy, we call this posterior merging as `Bayesian Arithmetic (BA)'. The above rule is commonly used in various distributed Bayesian computation, such as, Bayesian data fusion \citep{mutambara2019decentralized,durrant2001data,wu2022bayesian}, Bayesian Committee Machine \citep{tresp2000bayesian}, Consensus Monte Carlo \citep{scott2022bayes}, and approximate inference \citep{vehtari2020expectation,ashman2022partitioned}. 

\subsubsection{Merging as Posterior Correction}

Clearly, posterior merging does not recover the joint posterior over the whole data. We will now use posterior correction to write an exact expression for the error. Because both $q_i$ and $q_\llm$ are obtained by training a variational objetive, we can write their dual from as follows:
\[
   q_i \propto q_\llm \exp(-\losshat_{i})
   \qquad \text{and} \qquad
   q_\llm \propto \exp(-\losshat_{\llm}),
\]
where $\losshat_{i}$ is the site for $\loss_i$ evaluated at $q_i$ and $\losshat_{\llm}$ is the same for the whole pre-trained loss $\loss_\llm$ (we assume that it also contains the prior used during pre-training). The more corret notation would be to use $\losshat_{i|i}$, but we use $\losshat_i$ which is simpler.
Using a procedure similar to \cref{eq:postcorr}, we can rewrite the joint posterior shown in the second line,
\begin{equation}
   \begin{split}
      q_{\text{retrained}} &= \argmin_{q \in \mathcal{Q} } \,\, \myexpect_q \sqr{ \loss_\llm + \sum_{i=1}^t \alpha_i \loss_i } + \dkls{}{q}{p_0}\\
      &= \argmin_{q \in \mathcal{Q} } \,\, \dkls{}{q}{q_\ba} + \sum_{i=1}^t \alpha_i \underbrace{ \myexpect_{q}[\loss_i - \losshat_{i}] }_{\text{Task correction}} + \underbrace{ \myexpect_q[\loss_{\llm} - \losshat_{\llm}] }_{\text{LLM correction}} .
   \end{split}
   \label{eq:postcorr_merging}
\end{equation}
The result is obtained by simply multiplying and dividing by $q_\ba$ as we did in \cref{eq:postcorr}. The last two terms contain two kinds of corrections: the first one contains corrections to tackle the interference due to fine-tuned posteriors, while the second correction is needed to mitigate the interference to the original LLM. The reformulation above suggest to addressed both types of corrections to effectively merge the posteriors.

\subsubsection{Task Arithmetic and Hessian-Based Merging as Posterior Correction}
It is easy to check that the TA merging in \cref{eq:ta} can be recovered as a special case if we make two changes. First, we restrict the set of distributions to the $\qiso$ family, setting $q_\llm = \gauss(\vparam|\vm_\llm, \vI)$ and all $q_i = \gauss(\vparam|\vm_i, \vI)$ for $i=1,2,\ldots, t$. Second, we ignore the correction term. With these approximations, solving the above gives us $q_{\text{retrained}} = q_\ba$. The solution is then simply obtained by using
\cref{eq:ba} which yields $\vm_\ba = \vm_\llm + \sum_{i=1}^t \alpha_i (\vm_i - \vm_\llm)$, a merging equivalent to \cref{eq:ta} if we use the delta method to train both the LLM and perform fine-tuning. By ignoring the correction term, we obtain task arithmetic as a special instance of Bayesian arithmetic on the $\qiso$ family.

To get a better merging strategy, we can now explicitly approximate the correction terms, instead of ignoring it. We will use the same strategy as \cref{sec:if_pc} relying on the second-order Taylor expansion to approximate both the task and LLM corrections,
\begin{equation}
   \begin{split}
      \loss_i(\vparam) - \losshat_{i}(\vparam) &\approx \half (\vparam - \vm_i)^\top \vH_i (\vparam - \vm_i) \\
      \loss_\llm(\vparam) - \losshat_{\llm}(\vparam) &\approx \half (\vparam - \vm_\llm)^\top \vH_\llm (\vparam - \vm_\llm) 
   \end{split}
\end{equation}
where $\vH_i = \nabla^2 \loss_i(\vm_i)$ and $\vH_\llm = \nabla^2 \loss_\llm(\vm_\llm)$. Plugging these in \cref{eq:postcorr_merging}, we get the following estimate of the merged mean, as shown in \cref{app:deriv_merging},
\begin{equation}
   \vm_\ha = \vm_\llm + \sum_{i=1}^t \alpha_i \vH_\ha^{-1} (\vI + \vH_i) (\vm_i - \vm_\llm),
   \label{eq:ha}
\end{equation}
where $\vH_\ha = \vI + \vH_\llm + \sum_i \alpha_i \vH_i$.  
This is exactly the update derived by \citet[Eq. 12]{DaheimMPGK24} by using \emph{gradient mismatch}. In fact, it is easy to show that gradient mismatch is a special case of posterior correction. This is obtained by simply taking the derivatives of the correction term with respect to the mean $\vm$, 
\begin{equation}
   \nabla_\vm \myexpect_q[ \loss_i - \losshat_{i|t}] = \myexpect_q \sqr{ \nabla \loss_i } - \myexpect_{q_t} [\nabla \loss_i ] \,\, \approx\,\,  \nabla \loss_i(\vm) - \nabla \loss_i(\vm_t).
   \label{eq:grad_mismatch}
\end{equation}
The second equality follows by using Bonnet's theorem $\nabla_\vm \myexpect_q[\loss_i] = \myexpect_q[\nabla \loss_i]$ and the approximation afterward is due to the delta method. The correction term in this case account for gradient mismatches which can be seen as a special case of the natural-gradient mismatch used in this paper. Posterior correction generalizes their approach and relaxes the assumptions made by them (they assume a quadratic regularizer and no
such restrictions are needed in our derivation).

An alternate way to derive the same result is to expand the size of the family from $\qiso$ to $\qfull$, as we did in the previous section. Similar to the previous derivation, if we ignore the corrections, then merging reduces to the computation of $q_\ba$. Then, by denoting $q_i = \gauss(\vparam|\vm_i, \vS_i^{-1})$ and $q_\llm = \gauss(\vparam|\vm_\llm, \vS_\llm^{-1})$, we get the following merging formula, 
\begin{equation}
   \vm_\ba = \vm_\llm + \sum_{i=1}^t \alpha_i \vS_\ba^{-1} (\vI + \vS_i) (\vm_i - \vm_\llm),
\end{equation}
where $\vS_\ba = \vI + \vS_\llm + \sum_i \alpha_i \vS_i$. Essentially, the Hessians in \cref{eq:ha} are now replaced by the precision matrices. If we use the delta method to optimize over the $\qfull$ family, then this will reduce to Hessian-based merging. This again shows that we do not have to explicitly approximate the correction and a reduction naturally occurs when we use a more flexible posterior form, yielding a better adaptation method automatically. 


\subsection{Federated Learning as Posterior Correction}
\label{sec:fl}

The goal of distributed and federated learning \citep{mcmahan2016fedavg} is to learn a joint model $\vparam_\jnt$ at a server by communicating with local models $\vparam_i$ for $i=1,2, \ldots, t$ trained separately at $t$ clients. The local models use local losses $\loss_i$ not accessible at the server. This restriction forces the server to adapt all the available information and iteratively arrive at the correct solution.
We will now show that the methods used in federated learning can be seen as an iterative posterior-correction to eventually drive the corrections to zero. This is unlike the methods discussed so far which all used approximations for the correction term and therefore do not achieve perfect correction. Algorithms discussed here fix this issue.

\subsubsection{Federated Posterior Correction}

To show this, we start with a variational formulation for the joint-model learning.
For simplicity, we will assume that $q_\jnt$ and $p_0$ belong to the same exponential family because it simplifies the calculation, but the discussion is valid even when this condition does not hold. The variational formulation is shown below,
\begin{equation}
   \begin{split}
      q_\jnt^* &= \argmin_{q \in \mathcal{Q} } \,\, \myexpect_q \sqr{ \sum_{i=1}^t \loss_i } + \dkls{}{q}{p_0} \\
      &\,\,\propto p_0 \prod_{i=1}^t \exp\rnd{-\losshat_{i|\jnt}^*}, 
      \text{ where } \losshat_{i|\jnt}^*(\vparam) = \myang{\vT(\vparam), \natgrad \myexpect_{q_\jnt^*}[\loss_i]}.
   \end{split}
   \label{eq:fl_joint}
\end{equation}
In the second line, we write the solution in its dual form, similarly to \cref{eq:qdualform}.
The form factorizes across local models which suggests a natural adaptation strategy by exploiting the structure. Since each local \smash{$\losshat_{i|\jnt}^*$} contributes a factor, we can slowly try to move the locally learned approximations towards the desired \smash{$\losshat_{i|\jnt}^*$}. As we show, federated learning can be interpreted as following this recipe.

To show this, we first present a federated algorithm for posterior correction from which we can derive other federated learning algorithms. We denote by $\losshat_{i}$ a local site of $\loss_i$ at the local $q_i$. The algorithm starts with an initial value of \smash{$\losshat_{i} = \myang{\vT(\vparam), \dualparamhat_{i}}$} with a first guess for $\dualparamhat_{i}$. Then the algorithm follows a double loop structure. In the outer loop, we construct a joint posterior as follows,
\begin{equation}
   q_\jnt \gets \frac{1}{\mathcal{Z}_\jnt} p_0 \prod_{i=1}^t \exp\rnd{-\losshat_{i}}. 
      \label{eq:global_update}
\end{equation}
Then, we use it to write a posterior-corrected version of \cref{eq:fl_joint} (similarly to \cref{eq:postcorr}),
\begin{equation}
   q_\jnt^* = \argmin_{q \in \mathcal{Q} } \,\, \sum_{i=1}^t \myexpect_q \sqr{\loss_i - \losshat_{i} } + \dkls{}{q}{q_\jnt}. 
   \label{eq:bayes_fed}
\end{equation}
In the inner loop, we separately optimize with respect to each $q_i$ by solving 
\begin{equation}
   q_i \gets \argmin_{q_i \in \mathcal{Q} } \,\, \frac{1}{\rho} \myexpect_{q_i} \sqr{\loss_i - \losshat_{i} } +  \dkls{}{q_i}{q_\jnt}.
      \label{eq:local_update}
\end{equation}
Here, we multiply the local correction by $\rho$ as a simple approximation of the whole sum; for instance, we can set $\rho =1/t$ which is equivalent to weighting the sum by $t$. The site parameter is updated to be the natural gradients at the new $q_i$, that is, $\dualparamhat_{i} \gets \natgrad \myexpect_{q_i}[\loss_i]$. The outer and inner loops in \cref{eq:global_update} and \cref{eq:local_update} respectively are repeated until convergence. We call this algorithm the Federated Posterior Correction algorithm. It is easy to check that the solution in \cref{eq:fl_joint} is recovered upon
convergence; see \cref{app:fl_details}. As a result, upon convergence, the correction term also vanishes and $\losshat_{i}$ converge to $\losshat_{i|\jnt}^*$.

\subsubsection{Alternating Minimization Algorithm as Posterior Correction}

We will now show how existing federated learning algorithms can be seen as an implementation of the above federated posterior-correction algorithm. The closest algorithm to ours is perhaps the Alternating Minimization Algorithm (AMA) of \citet{Ts91} which uses splitting methods for optimization. For instance, consider the following objective 
\begin{equation}
   \vparam_\jnt^* = \argmin_{\vparam} \sum_{i=1}^t \loss_i(\vparam) + \half \|\vparam\|^2.
   \label{eq:federateMAP}
\end{equation}
A quadratic regularizer is used above for simplicity but the method can be used for a much larger class of objectives.

A detailed description of AMA is given in \cref{app:ama}. Below, we summarize the updates and then compare them to our federated algorithm.
In AMA, we define local loss functions $\loss_i(\vparam_i)$ over local parameters $\vparam_i$ for all $i=1, 2,\ldots, t$. A global parameter $\vparam_\jnt$ is also introduced which operates over the quadratic regularizer $\|\vparam_\jnt\|^2/2$. Then a Lagrangian is formed to enforce constraints $\vparam_i = \vparam_\jnt$ for all $i$ via \emph{dual} vectors
$\vv_i$ (also known as Lagrange multipliers). 
The Lagrangian gives rise to the following update for the triplet $(\vparam_\jnt, \vparam_i, \vv_i)$,
\begin{equation}
   \vparam_\jnt \gets -\sum_{i=1}^t \vv_i,
   \quad\quad
   \vparam_i \gets \argmin_{\vparam_i} \,\, \loss_i(\vparam_i) - \vparam_i^\top \vv_i + \rho \half \|\vparam_i - \hat\vparam_\jnt \|^2,
   \quad\quad
   \vv_i \gets \nabla \loss_i(\vparam_i) .
   \label{eq:ama_update}
\end{equation}
To show the similarity to our federated algorithm, we rewrite \cref{eq:global_update,eq:local_update} in a similar form where we update the triplet $(\vnatparam_\jnt, \vmeanparam_i, \dualparamhat_i)$,
\begin{equation}
   \vnatparam_\jnt \gets - \sum_{i=0}^t \dualparamhat_i,
   \quad\quad
   \vmeanparam_i \gets \argmin_{\vmeanparam_i} \myexpect_{q_i}[\loss_i] - \myang{\vmeanparam_i, \dualparamhat_i} + \rho \dkls{}{q_i}{q_\jnt} ,
   \quad
   \dualparamhat_i \gets \natgrad \myexpect_{q_i}[\loss_i] .
   \label{eq:2stagePoCoNatparamForm}
\end{equation}
A derivation is in \cref{app:amaaspc}.  Essentially, the global natural parameter $\vnatparam_\jnt$ is updated by summing the local site parameters $\dualparamhat_i$, which are the natural gradients of the expected local-loss functions. The role of the dual vector $\vv_i$ is now played by the site parameters $\dualparamhat_i$ (a minor detail is that $\dualparamhat_0$ is the natural parameter of $p_0$). The second update is performed in the $\vmu_i$-space where the linear term $\vparam^\top\vv_i$ is now replaced by the $\myang{\vmeanparam_i, \dualparamhat_i}$.
The theorem below shows AMA can be recovered as a special case. 
\begin{thm}
   For $\qiso$ family, \cref{eq:2stagePoCoNatparamForm} reduces to \cref{eq:ama_update} if we approximate $\myexpect_{q_i}[\loss_i] \approx \loss_i(\vm_i)$.
\end{thm}
A detailed proof is given in \cref{app:amaaspc}. Overall, these results show that not only can we recover AMA from our algorithm, we can also generalize it by choosing more flexible exponential-family forms.

\subsubsection{Partition Variational Inference as Posterior Correction}

Our algorithms closely resembles Partitioned Variational Inference (PVI) \citep{ashman2022partitioned}. This also highlights some potential issues with the PVI algorithm and provides ways to fix them. PVI aims for distributed federated Bayesian inference where the losses $\loss_i$ are simply negative of log-likelihoods, that is, $\log\lik_i = - \loss_i$. 
PVI follows the strategy of the well-known Expectation Propagation (EP) algorithm, where local site functions, denoted by $\site_i$, are used to approximate the likelihood, that is, $\site_i \approx \lik_i$. A major difference is that the PVI algorithm does not assume that $q$ is an EF posterior, so $\site_i$ may not necessarily be defined in terms of natural gradients. In such cases, PVI implements the following steps, starting with $q_\jnt \gets p_0$ and $\site_i \gets 1$,
\begin{equation}
   \begin{split}
      &q_i \gets \argmin_{q_i \in \mathcal{Q} } \,\, \myexpect_{q_i} \sqr{\loss_i + \log \site_i} +  \dkls{}{q_i}{q_\jnt} \\
      &\site_i^\old \gets \site_i,\,\, \text{ and } \,\, \site_i \gets \site_i^\old \frac{q_i}{q_\jnt}\\
      &q_\jnt \gets \frac{1}{\mathcal{Z}_\jnt} q_\jnt \prod_{i=1}^t \frac{\site_i}{\site_i^\old}. 
   \end{split}
   \label{eq:pvi}
\end{equation}
We can show that, for EF posteriors, these three steps equivalent to \cref{eq:2stagePoCoNatparamForm} if we set $\rho=1$. For such cases, the sites are simply \smash{$\site_i \propto \exp(-\losshat_{i})$}. A detailed derivation of the equivalence is in \cref{app:pvi_deriv}.

The choice $\rho=1$ in PVI may not be appropriate, as also noted by \cite{ashman2022partitioned}, the algorithm can diverge without additional ``damping''. An appropriate range of $\rho$ which ensures convergence is discussed in \citet{Ts91}. For instance, when using a quadratic regularizer as in \cref{eq:federateMAP}, the range $\rho \in(0, 2/t)$ ensures convergence under mild assumptions on $\loss_i$. Clearly, for $t >2$, this range never includes $\rho=1$.
To improve stability, \citet{ashman2022partitioned} introduced a damping factor $\rho\in(0,1)$ to update $\site_i$ but this is not used in the update of $q_i$. It is unclear whether the resulting procedure converges or not, but it does slow down the update; see further discussion in \cref{app:pvi_deriv}. In practice, we do observe cases where PVI can diverge \citep{moellenhoff2025} and recommend a careful use of $\rho$ by using the suggestions of \citet{Ts91}.


\subsubsection{ADMM-Style Extensions of Federated Posterior Correction}

It is also possible to use updating schemes for posterior corrections that are similar to Alternating Direction Method of Multipliers (ADMM) \citep{GaMe76,GlMa75}. A major difference between ADMM and AMA is in the former's update of $\vparam_\jnt$ where an additional sum over $\vparam_i$ is included, as shown below,
\begin{equation}
   \vparam_\jnt \gets \alpha \sum_{i=1}^K \vv_i + (1-\alpha) \sum_{i=1}^K \vparam_i ,
\end{equation}
where $\alpha = 1/(1+ \rho K)$. The federated posterior correction can easily be modified to include such updating, for instance, we can replace the update of $q_\jnt$ in \cref{eq:global_update} by the following:
\begin{equation}
   q_\jnt \gets \frac{1}{\mathcal{Z}_\jnt} \sqr{p_0 \prod_{i=1}^t \exp\rnd{-\losshat_{i}}}^{\alpha} \sqr{\prod_{i=1}^t q_i }^{1-\alpha} .
   \label{eq:admm}
\end{equation}
This update can also be interpreted as a BLR step at $\prod_i q_i$ with learning rate $\alpha$. A similar extension of the BLR is recently proposed by \citet{moellenhoff2025} who also use the dual form of the VB solution, referred to as the \emph{Bayesian Duality} structure. They are the first to derive both AMA and ADMM as special cases of a federated Bayesian algorithm based on such a dual structure. They also propose a modification of PVI, as well as a new variant of the
algorithm by \citet{SwKh25},
and present their ADMM variants, such as, FedADMM \citep{gong2022fedadmm,wang2022fedadmm,ZhLi23} and FedDyn \citep{acar2021feddyn}.
These extensions bring new insight into the existing Bayesian literature \citep{vehtari2020expectation,ashman2022partitioned} which is currently missing a rigorous connection to decade old work in convex duality and distributed optimization.
Altogether, with this type of connections, posterior correction opens a new way to design better adaptive algorithms.



\section{Method Details and Further Connections}

In this section, we give further details on the dual perspective of the Bayesian Learning Rule (BLR), as well as alternate interpretations of posterior correction. Then, we discuss a few more details regarding the connections to continual learning methods. Specifically, we derive expressions for correction terms, derive K-priors as a special case, and discuss application to online variational inference. 

\subsection{A Dual Perspective of The Bayesian Learning Rule (BLR)}
\label{sec:dualBLR}

Posterior correction is derived by using a dual perspective of the Bayesian Learning Rule (BLR) by \citet{khanRue23}. The duality is similar to those used in the optimization literature, for example, the closed-circuit of \citet[Fig. 2]{Ro67}. Such perspectives do have roots in the Bayesian literature, for example, the representer theorem by \citet{kimeldorf1970correspondence} was original proposed in a Bayesian context. Unfortunately, these ideas are rarely used in the modern Bayesian
machine-learning community, but they remains popular in the kernel-methods community \citep{scholkopf2001generalized}. The dual-perspective of the BLR connects them back to Bayes via the variational-Bayesian route. 

In this section, we give additional details about the dual form shown in \cref{eq:qdualform}. The form represents the solution $q_t$ as a product of multiple factors, traditionally referred to as `sites' in the graphical-model literature. The expression is derived by using the expectation parameters $\vmeanparam = \myexpect_q[\vT(\vparam)]$. The pair $(\vnatparam, \vmeanparam)$ live in two spaces dual to each other and connected by a bijective Legendre transform \citep[Sec 2.2]{khanRue23}.
As a consequence, the natural-gradient in the $\vnatparam$-space are simply gradients in the $\vmu$-space, that is, $\natgrad_{\vnatparam} = \nabla_{\vmeanparam}$ \citep[Thm. 1]{khan2018fast1}. This fact is used in the BLR to simplify the computation of natural gradients.

With this, the dual form in \cref{eq:qdualform} is obtained by applying \smash{$\natgrad$} to \cref{eq:ERMasVL} and setting to 0. The calculation is shown below where we rewrite $\dkls{}{q_t}{p_0} = \myexpect_{q_t}[\loss_0] - \entropy(q_t)$ where $\entropy(q_t)$ is the entropy, then get the second equality by using the fact that \smash{$\natgrad\entropy (q_t) = -\vlambda_t$}, and write the third equation by substituting $\vnatparam_t$ in $q_t \propto
\exp(\myang{\vT(\vparam), \vnatparam_t})$:
\begin{equation}
   \sum_{i=0}^t \natgrad \myexpect_{q_t} \sqr{\loss_i} - \natgrad \entropy(q_t) = 0 
   \,\,\,\implies\,\, 
   \vlambda_t = \sum_{i=0}^t \natgrad \myexpect_{q_t}[-\loss_i] 
   \,\,\,\implies\,\, 
   q_t \propto \prod_{i=0}^t e^{-\myang{\vT(\vparam), \natgrad \myexpect_{q_t}[\loss_i] } } .
   \label{eq:deriv_dual}
\end{equation}
The derivation in \cref{eq:deriv_dual} reveals the underlying dual structure: The left side in the second expression belongs to the $\vlambda$ space while the gradients in the right are computed in the dual $\vmu$ space. This is akin to the dual structure used in \citet[Fig. 2]{Ro67} and other forms of representer theorems. A formal derivation can be done via a Lagrangian formulation which shows that natural gradients can also be seen as Lagrange multipliers. 
We skip this discussion as it is not our main focus in this paper, but interested readers may see similar formulations in \citet{KhAr13,adam2021dual,moellenhoff2025}. The extension to distributions with non-constant base measure is discussed in \cref{app:base_measure}.

The dual form also holds during learning when using the BLR to optimize the variational objective. The BLR uses the following moving average with a learning rate $\rho$, which in a similar fashion gives rise to a dual form that holds during training,
\begin{equation}
   \vlambda_t \leftarrow (1-\rho) \vlambda_t + \rho \sum_{i=0}^t \natgrad \myexpect_{q_t}[-\loss_i] 
   \qquad\implies \qquad
   q_t \leftarrow \frac{1}{\mathcal{Z}_t} q_t^{1-\rho}\prod_{i=0}^t \exp\rnd{- \rho\losshat_{i|t} } .
   \label{eq:deriv_dual}
\end{equation}
A logistic regression example is included in \cref{app:dual_training}.
One can go even further and write a dual form by `unrolling' all the updates to an older iteration, say $j$ step back in the past. This is similar to the `lazy' form used in online learning \citep{hoeven2018many} and was originally proposed by \cite{khan2017conjugate}; see also \citet[Eq. 60]{khanRue23}. 

We will now derive the explicit form of the site function for Gaussian distributions shown in \cref{tab:site}. For such cases, the sites take a similar form to surrogates obtained by using Taylor's method. We first consider the $\qiso_t = \gauss(\vparam|\vm_t, \vI)$ for which we have $\vT(\vparam) = \vparam$ and $\vmu_t = \myexpect_{q_t}[\vparam] = \vm_t$. Using this, we can write
\begin{equation}
   \losshatiso_{i|t}(\vparam) = \myang{ \vparam, \nabla_{\vm} \myexpect_{q_t}[\loss_i]} =  \vparam^\top \myexpect_{q_t}[\nabla \loss_i],
\end{equation}
where in the last expression we push the gradient inside the expectation by using the Bonnet's theorem \citep{bonnet1964transformations, khanRue23}. We can also write a similar expression for a $\qfull = \gauss(\vparam|\vm_t, \vS_t^{-1})$ for which we have $\vT(\vparam) = (\vparam, \vparam\vparam^\top)$. Using this, we get
\begin{equation}
   \begin{split}
      \losshatfull_{i|t}(\vparam) &= \myang{ \vT(\vparam), \natgrad \myexpect_{q_t}[\loss_i] } \\
      & = \vparam^\top \myexpect_{q_t}\sqr{ \nabla \loss_i - (\nabla^2 \loss_i) \vm_t} + \half \vparam^\top \myexpect_{q_t}\sqr{\nabla^2 \loss_i} \vparam \\
      & = \vparam^\top \myexpect_{q_t}[\nabla \loss_i] + \half (\vparam-\vm_t)^\top \myexpect_{q_t}\sqr{\nabla^2 \loss_i} (\vparam-\vm_t) + \cnst
   \end{split}
\end{equation}
where in the first line we make use of the expression given in \citet[Eq. 11]{khanRue23}. In the derivations above, we have ignored the constants but these expressions are essentially a result of the $\first$-order Taylor expansion in the (lifted) $\vmu$-space used in the BLR during optimization. For instance, denoting $L_i(\vmeanparam) = \myexpect_q[\loss_i]$, we apply a mirror-descent update which linearizes $L_i$ in the $\vmu$-space,
\begin{equation}
   \begin{split}
      \vnatparam_t &\gets \arg\min_{\vmeanparam} \sum_{i=0}^t \sqr{ L_i(\vmeanparam_t) + \myang{ \vmu -\vmu_t, \grad_{\vmeanparam} L_i(\vmeanparam_t)} } - \entropy(q_t) + \frac{1}{\rho}\dkls{}{q}{q_t} \\
      &= \arg\min_{\vmeanparam} \sum_{i=0}^t \myexpect_{q} \myang{ \vmeanparam, \natgrad \myexpect_{q_t}[\loss_i] } + \frac{1}{\rho}\dkls{}{q}{q_t^{1-\rho}},
   \end{split}
\end{equation}
where we ignored the constants and absorbed the entropy term in the KL term. This form is originally proposed in \citet{khan2017conjugate}. Information geometry based explanations are given in \citet{khan2025information}.

\subsection{Posterior Correction as Bayes' Filter and Natural-Gradient Mismatch}
\label{sec:postcorr_details}

The posterior correction in \cref{eq:postcorr} is derived using the variational form but the update can also be written in a form similar to Bayes' rule. Essentially, we apply the dual form given in \cref{eq:qdualform} to \cref{eq:postcorr} which gives the following form for $q_{t+1}$ in terms of $q_t$,
\begin{equation}
   \begin{split}
      q_{t+1} = \frac{\mathcal{Z}_t}{\mathcal{Z}_{t+1}} \rnd{ e^{-\losshat_{t+1|t+1}} } \times q_t \times \prod_{i=0}^{t}  \rnd{ e^{-\rnd{\losshat_{i|t+1} - \losshat_{i|t}}} }
   \end{split}
   \label{eq:var_recurse_corr}
\end{equation}
The recursive form is similar to Bayesian filtering but there is an additional third term to account for the interference introduced in the past $\loss_i$ due to the inclusion of $\loss_{t+1}$ (a similar update for federated learning is in \cref{eq:bayes_fed}). Using this, we can also write the divergence to $q_{t+1}$ from $q_t$ by simply rearranging \cref{eq:var_recurse_corr} to write $\myexpect_{q_{t+1}}[\log (q_{t+1}/q_t)]$ as follows,
\begin{equation}
   \dkls{}{q_{t+1}}{q_t} = \myexpect_{q_{t+1}} \sqr{ -\losshat_{t+1|t+1} } - \sum_{i=0}^t \myexpect_{q_t}\sqr{  \losshat_{i|t+1} - \losshat_{i|t} } + \log \frac{\mathcal{Z}_t}{\mathcal{Z}_{t+1}}.
   \label{eq:knowledge_gap}
\end{equation}
The last two terms account for the normalizing constants of the two distributions.

This expression can be seen as an extension of the Information Gain \citep{lindley1956measure}, which is defined as $\dkls{}{p_{t+1}}{p_t} = \myexpect_{p_{t+1}}[-\loss_{t+1}] + \log (\mathcal{Z}_t/\mathcal{Z}_{t+1})$ where $p_t$ and $p_{t+1}$ are the old and new posteriors respectively. The first term in \cref{eq:knowledge_gap} is similar but uses the site instead of the loss. The second
term is additional and arises due to the correction. To the best of our knowledge, no such closed-form expression exist to the date to characterize information gain of variational posteriors. For smaller gains, adaptation should be quicker. This is how posterior correction quantifies the feasibility of quick adaptation.

Another view of posterior correction is to see it as the natural-gradient mismatch. This is obtained by simply writing the natural parameters $\vlambda_{t+1}$ of $q_{t+1}$ in \cref{eq:var_recurse_corr},
\begin{equation}
   \vlambda_{t+1} = \natgrad \myexpect_{q_{t+1}}[-\loss_{t+1}] + \vlambda_t - \sum_{i=0}^{t}  \underbrace{ \rnd{  \natgrad \myexpect_{q_{t+1}}[\loss_i] - \natgrad \myexpect_{q_{t}}[\loss_i] } }_{\text{Mismatch}}.
   \label{eq:nat_grad_mismatch}
\end{equation}
The mismatch is yet another characterization of the interference. Posterior correction essentially replaces the ``stale'' natural gradients by the fresh new ones. For example, the PVI algorithm discussed in \cref{sec:fl} implements this exact operation for posterior correction.

Finally, we discuss how to apply posterior correction during training. We consider a simple case where we want to boost the training of another model given a `check-point' with a dual form
\[
   q_{\chk} \propto p_0 \prod_{i=1}^t \exp(- \losshat_{i|\chk}) .
\]
For simplicity, we assume that the training data for the checkpoint is the same as the training which we want to boost. We also assume that it involves the same prior $p_0$ which takes the same exponential form as the model. However, we note that the procedure described below works for much more general cases. In fact, we can also have multiple check points stored during training construct a Bayesian Arithmetic average, for instance, as shown in \cref{eq:ba} and \cref{eq:admm}. The case
below is just chosen for the simplicity sake.

We can use $q_\chk$ to boost the training trajectories of the BLR where we simply replace the prior $p_0$ by the $q_\chk$ and correct the losses $\loss_i$ accordingly. This is shown below where we simplify the BLR update of \citet[Eq. 22]{khanRue23} taken at $q_\old$,
\begin{equation}
   \begin{split}
      q_\new &=  \argmin_{\vmu} \,\, \vmu^\top \natgrad \rnd{ \sum_{i=1}^t  \myexpect_{q_{\vmu}}[\loss_i] + \dkls{}{q_{\vmu}}{p_0} } + \frac{1}{\rho}\dkls{}{q_{\vmu}}{q_\old} \\
      &=  \argmin_{\vmu} \,\,  \sum_{i=1}^t \vmu^\top \natgrad \myexpect_{q_{\vmu}}[\loss_i] + \frac{1}{\alpha}\dkls{}{q_{\vmu}}{ p_0^\alpha q_\old^{1-\alpha} } \\
      &=  \argmin_{\vmu} \,\,  \sum_{i=1}^t \vmu^\top \natgrad \myexpect_{q_{\vmu}}[\loss_i - \losshat_{i|\chk}] + \frac{1}{\alpha}\dkls{}{q_{\vmu}}{ {q_\chk}^\alpha q_\old^{1-\alpha} }. 
   \end{split}
\end{equation}
The first line is obtained by simply plugging in the VL objective. The second line is simplified by noting that the KL between $q_{\vmu}$ and $p_0$ contains conjugate term so it can be simply taken out and merged with the last KL term by redefining $\alpha = \rho/(1+\rho)$. An update of this form is in \citet[Eq. 59]{khanRue23} and a derivation for the conjugate prior case is given in \citet[Eq. 26]{nickl2023memory}. The final step is obtained by also noting the fact
that $\losshat_{i|\chk}$ is also a conjugate form, so it can simply be inserted inside the first term.

The derivation above shows the strength of the posterior correction. Not only that it can be applied to trained model, it can also be used to boost the training. The check-point also need not belong to a trained model. As long as we can express them in a dual form, we can use their stored knowledge to boost training of other models. This particular feature makes posterior correction a powerful mechanism to reuse and repurpose the knowledge learned during model training.

\subsection{Posterior Correction for Continual Learning}
\label{sec:cl_corr_term}

We start with the derivation of \cref{eq:pred_match_general} which shows that the correction term for continual learning can be written as a sum of two terms. The loss function considered is a Bregman divergence (or equivalently using an exponential family) which takes the following form,
\[
   \glmloss[y_i, \hat y_i (\vparam)] = - y_i f_i(\vparam) + A(f_i(\vparam)) 
\]
where $A(\cdot)$ is the convex function that generates the Bregman divergence. Typical examples include cross-entropy loss which commonly in multi-class classification with neural networks. There, the model output (also called logits) are vectors of $f_i^k(\vparam)$ for each class $k$, the function $A(\cdot)$ is $\log\sum_{k=1}^K \exp(f_i^k(\vparam))$, the log-sum-exp function over all $K$ classes. The predictions are obtained by simply taking a softmax over $f_i^k(\vparam)$.
For simplicity, we will present the derivation for a scalar $f_i(\vparam)$.

For such loss functions, the correction term can be simplified as follows,
\begin{equation}
   \begin{split}
      \loss_i(\vparam) - \losshatiso_{i|t}(\vparam) &= \glmloss[y_i, \hat y_i (\vparam)] - \vparam^\top \myexpect_{q_t} \sqr{ \nabla \glmloss[y_i, \hat y_i(\vparam) ] } \\
      &= - y_i f_i(\vparam) + A(f_i(\vparam)) - \vparam^\top \myexpect_{q_t} \sqr{ \nabla f_i(\vparam) \rnd{ \hat y_i(\vparam) - y_i} } \\
      &= - \hat y_{i|t} f_i(\vparam) + (\hat y_{i|t} - y_i) f_i(\vparam) + A(f_i(\vparam)) - \vparam^\top \myexpect_{q_t} \sqr{ \nabla f_i(\vparam) \rnd{ \hat y_i(\vparam) - y_i} } \\
      &= - \hat y_{i|t} f_i(\vparam) + r_{i|t} f_i(\vparam) + A(f_i(\vparam)) - \vparam^\top \myexpect_{q_t} \sqr{ \nabla f_i(\vparam) \rnd{ \hat y_i(\vparam) - y_i} } \\
      &\approx - \hat y_{i|t} f_i(\vparam) + r_{i|t} f_i(\vparam) + A(f_i(\vparam)) - \vparam^\top { \nabla f_i(\vm_t) \rnd{ \myexpect_{q_t} [\hat y_i(\vparam)] - y_i} } \\
      &= - \hat y_{i|t} f_i(\vparam) + A(f_i(\vparam)) + r_{i|t} f_i(\vparam) - r_{i|t} \vparam^\top  \nabla f_i(\vm_t) \\
      &= \glmloss\sqr{\hat{y}_{i|t}, \, \hat{y}_i(\vparam) } + r_{i|t} [  f_i(\vparam) - \hat{f}^{\text{lin}}_i(\vparam) ] + \cnst
   \end{split}
   \label{eq:pred_match_general_deriv}
\end{equation}
The first and second line follows from the definition of the loss and site. In the third line, we subtract and add $\hat y_{i|t} f_i(\vparam)$ where $\hat y_{i|t} = \myexpect_{q_t}[ \hat{y}_i(\vparam)]$. Then, we write it in terms of the residual $r_{i|t} = \hat y_{i|t} - y_i$ in the fourth line. In the fifth line, we use the $\first$-order delta method in the last term approximating $\nabla f_i(\vparam) \approx \nabla f_i(\vm_t)$ and move the expectation with respect to $q_t$
inside over to $\hat y_i(\vparam)$. The last two lines are simple rearrangements and written in terms of $\hat f^{\text{lin}}_i(\vparam) = (\vparam - \vm_t)^\top \nabla f_i(\vm_t)$ (with a term that is constant with
respect to $\vparam$). The only assumption made is the delta method for the expectation which might be useful for Bayesian cases but otherwise can be safely ignored. Other than that, the expression holds for general models, such as, neural networks.

The derivation directly extends to $\qfull$, where we get an additional quadratic term, 
\begin{equation}
   \begin{split}
      \loss_i(\vparam) - \losshatfull_{i|t}(\vparam) &= \loss_i(\vparam) - \losshatiso_{i|t}(\vparam) -\half (\vparam-\vm_t)^\top \vH_{i|t} (\vparam -\vm_t) \\
      &\approx \glmloss\sqr{\hat{y}_{i|t}, \, \hat{y}_i(\vparam) } + r_{i|t}  \sqr{ f_i(\vparam) - \hat{f}^{\text{quad}}_i(\vparam) } - \half \beta_{i|t} \| \nabla f_i^{\text{lin}}\|^2 ,
   \end{split}
   \label{eq:corr_qfull}
\end{equation}
where we denote $\hat f_i^{\text{quad}}(\vparam) = \hat f_i^{\text{lin}}(\vparam) + \half (\vparam-\vm_t)^\top \nabla^2 f_i(\vm_t) (\vparam-\vm_t)$, and $\beta_{i|t} = \myexpect_{q_t}[A''(f_i(\vparam)) ]$.
A derivation is in \cref{app:qfull_corr_cl} which involves writing the Hessian of the loss in terms of the Generalized Gauss-Newton (GGN) matrix and the Hessian of the model output. Applying the delta method, then gives this approximation. The linearization error is now reduced due to the use of a better quadratic surrogate, as opposed to the linear surrogate used in \cref{eq:pred_match_general_deriv}. The last term further reduces the error and is due to the GGN matrix.

We discuss some examples to illustrate the form of the correction. For linear regression, as we saw before in \cref{eq:corr_linreg}, that for the $\qiso$ family the correction is simply the prediction mismatch, but for the $\qfull$ the correction entirely vanishes due to the GGN term, 
\[
   \loss_i(\vparam) - \losshatfull_{i|t}(\vparam) = \| \vx_i^\top\vparam - \vx_i^\top \vparam_t\|^2 - \half \|(\vparam - \vm_t)^\top \vx_i \|^2 = 0.
\]
This makes sense because a full Gaussian posterior is the exact posterior and there is no need for a correction. Therefore, posterior correction simply reduces to Bayes' rule. Consider another example for logistic regression where $\hat y_i(\vparam) = \sigmoid(\vx_i^\top \vparam)$ with $\sigma(\cdot)$ denoting the sigmoid function. The $\qiso$ family yields the usual prediction matching term but the $\qfull$ family reduces this by comparing the linear outputs,
\begin{equation}
   \begin{split}
      \loss_i(\vparam) - \losshatfull_{i|t}(\vparam) = \glmloss \sqr{ \sigmoid(\vx_i^\top\vparam), \sigmoid(\vx_i^\top \vparam_t) } - \half \beta_{i|t} \|\vx_i^\top\vparam - \vx_i^\top\vm_t \|^2 
   \end{split}
\end{equation}
The second term essentially attempts to improve the current posterior $q_t$ by removing the stale $\beta_{i|t}$ and replace it by the fresh ones by using the prediction mismatch term. By using a more flexible posterior, we reduce the correction required for accurate adaptation. This idea was first used in K-priors by \citet[Fig. 3a]{khan2021knowledge} through gradient matching but it is also connected to PVI where we replace old surrogates by new ones during federated learning.
In general, all such procedures are generalized via posterior correction. We next show that the correction term generalizes K-priors to exponential-family posteriors.

\subsection{Knowledge-Adaptation Priors as Posterior Correction}
\label{sec:kprior_details}

The prior shown in \cref{eq:varkprior} is a generalization of the K-prior by \citet{khan2021knowledge}. We will now show this for the K-prior presented in \citet[Eq. 8]{khan2021knowledge} for a linear model $f_i(\vparam) = \vx_i^\top \vparam$ over the Bregman loss $\loss_i(\vparam) = \glmloss(y_i, \hat y_i(\vparam))$ and quadratic regularizer $\loss_0(\vparam) = \half \delta \|\vparam\|^2$. We have already derived the correction terms for $i=1,2,\ldots,t$. Therefore, we just need the correction term for
$\loss_0$ and the KL term. We derive this below.


The correction term for $\loss_0$ in the case of $\qiso$ takes a bit more effort because the base measure is not constant: $h(\vparam) = -\half \|\vparam\|^2 - \half P\log (2\pi)$. For such a case, we use 
\begin{equation}
   \begin{split}
      \loss_0(\vparam) - \losshatiso_{0|t}(\vparam) &= \loss_0(\vparam) - \vparam^\top \myexpect_{q_t}[ \nabla \loss_0 + \nabla \log h] + \log h(\vparam) \\
      &= \half \delta \|\vparam\|^2 - \delta \vparam^\top\vm_t + \vparam^\top\vm_t - \half \|\vparam\|^2 \\
      &= \half \delta\|\vparam-\vm_t\|^2 - \half \|\vparam - \vm_t\|^2 + \cnst \\
   \end{split}
   \label{eq:kprior_deriv_l0}
\end{equation}
Adding this to the KL term, we get
\begin{equation}
   \myexpect_q[\loss_0 - \losshat_{0|t}] + \dkls{}{q}{q_t} = \half\delta\|\vm-\vm_t\|^2 + \cnst
   \label{eq:corr_l0}
\end{equation}
Substituting in \cref{eq:varkprior}, we obtain 
\begin{equation}
   \kprior_t^{\text{iso}}(\vm) = \sum_{i=1}^t \myexpect_q \rnd{ \glmloss\sqr{ \hat y_{i|t},\, \hat y_i(\vparam) } } + \delta \half \|\vm - \vm_t\|^2 
\end{equation}
which is exactly \citet[Eq. 8]{khan2021knowledge} with respect to $\vm$, if we use the delta method: $\hat y_{i|t} = \myexpect_{q_t} [ \hat y_i(\vparam)] \approx \hat y_i(\vm_t)$ and $\myexpect_q \rnd{ \glmloss\sqr{ \hat y_i(\vm_t),\, \hat y_i(\vparam) } } \approx \glmloss[ \hat y_i(\vm_t),\, \hat y_i(\vm) ] $.

The prior in \cref{eq:varkprior} extends K-priors to generic exponential-family posterior forms. We will call this prior the \emph{Variational} K-prior. The new family of K-priors extends the prediction matching idea to other type of matching. For instance, the derivatives of the correction term in \cref{eq:varkprior} can be written as gradient mismatch as shown in \cref{eq:grad_mismatch}.
Similarly, if we use $\qfull$, the correction terms implement both gradient and Hessian matching. In general, such matching of predictions, gradients, or Hessians naturally emerges through natural-gradient mismatch in \cref{eq:nat_grad_mismatch}. All we need to do is to choose an appropriate family to match the desired natural gradients.

The general form of variational K-prior also makes it easy to mix and match different regularization methods. For instance, using the correction term expression given in \cref{eq:corr_qfull} for $\qfull$ family, we can make a decision to store specific examples for memory replay. For instance, examples whose residuals $r_{i|t}$ are high can accumulate errors through the nonlinear term, so it is better to include them through replay. For the other examples, it might be enough to simply
store a representation of the inputs. Such a mixture would give rise to a memory set where we pick a set of examples for replay $\memory_{\text{rep}}$ and then another \emph{disjoint} set for prediction matching $\memory_{\text{pred}}$, and use the following correction term,    
\[
   \sum_{i\in \memory_{\text{rep}}} \rnd{ \loss_i - \losshat_{i|t} } +  \sum_{j\in {\memory_{pred} } } \rnd{  \glmloss\sqr{\hat{y}_{j|t}, \, \hat{y}_j(\vparam) } - \half \beta_{i|t} \| \nabla f_i^{\text{lin}}\|^2 }.
\]
The two sets have to be disjoint so as to not double count the contribution of prediction matching. The benefit of using the second term above is that those examples do not need labels and they can be summarized using arbitrary input locations, for instance, similar to a core-set or set of inducing inputs. \citet{daxberger2023improving} used a memory construction similarly to the above for K-prior to get good improvement on the ImageNet data set. Similarly, \citet{pan2020continual} and
\citet{khan2021knowledge} used a quantity similar to $\beta_{i|t}$ to pick examples to do prediction matching on. They all show consistent improvement and point this as a viable option to design better memory replay methods.

We now briefly discuss connections to other approaches covered under the K-prior framework. For example Knowledge Distillation (KD) \citep{hinton2015distilling} considers a teacher-student learning scenario which can also be written as a posterior correction. The KD objective uses a convex combination for $\gamma \in (0,1)$ which can be rewritten as,
\begin{align*}
   \gamma \sum_{i=1}^t \glmloss \sqr{ \hat y_i(\vparam_t ), \,  \hat y_i(\vparam) } +  (1-\gamma)\sum_{i=1}^t \glmloss \sqr{ \hat y_i(\vparam_t ), \,  \hat y_i(\vparam) } 
   = \sum_{i=1}^t \glmloss \sqr{ \hat y_i(\vparam_t ), \,  \hat y_i(\vparam) } + \gamma \, \sum_{i=1}^t r_i(\vparam_t) f_i(\vparam),
\end{align*}
where $r_i(\vparam)$ are the residuals. Essentially, the parameter $\gamma$ is used to down-weight the mistakes made by the teacher. The objective shares similarities to \cref{eq:pred_match_general_deriv} where we explicitly `correct' the mistake not just down-weight them. In KD, we have the luxury to train on all examples, but it is not always possible to store the whole training set. Posterior correction can handle such cases where we need to pay attention to specific mistakes of the teacher and
ensure they do not get transferred to the teacher. Note that, typically, the architectures or teacher and students are different, so one need to choose an appropriate parameterization and divergence function to define a valid KL term. We do not go into details of this since this is out of scope for our work. 

Posterior correction is also closely related to dual approaches in SVM, such as, incremental Support Vector Machines \citep{cauwenberghs2001incremental, liang2009incremental}, Similarity Control \citep{vapnik2015learning}, and their extensions to neural networks, for example, \citep{lopez2017gradient}. The correction term can be written in a trust-region form with constraints $\myexpect_q[\loss_i] = \myexpect_q[\losshat_i]$. The dual problem can be
connected to posterior correction via a Lagrangian. The continual learning case can be written in a similar fashion.

\subsection{Posterior Correction for Sequential and Online Variational Inference}


It is well-known that the recursive elegance of Bayes' rule is lost when approximations are used. The problem of an incorrect $\hat q_{t+1}$ when \cref{eq:vcl} is used occurs in almost all approximate inference problems. This is true for even the simplest variational (Bayesian) learning on conjugate models where mean-field approximations are used \citep{sato2001online, ghahramani2000online, honkela2003line, hoffman2010online,
wang2011online}, but also more recent variants of those \citet{broderick2013streaming, cuong2018, zeno2018task, cherief2019generalization, jones2024bayesian}. Other approximations also suffer from this issue, for example, those proposed in \citet{opper1999bayesian,
winther1998optimal, Heskes2002ExpectationPF, csato2002sparse}, among many others. 
We expect similar problems to exist even when amortized inference is used \citep{archer2015black, kim2020variational, krishnan2017structured, campbell2021online}.

The posterior correction is a useful approach to fix such issues in sequential and online variational inference. As an example, we show that an existing method of \citet{bui2017streaming} is in fact an example of posterior correction for Sparse Variational Gaussian Process (SVGP). The method attempts to handle a challenging case of updating both the inducing input as well as hyperparameters but we will consider a simpler version to show similarities to posterior correction. 
SVGP uses a set of inducing variables, denoted by $\vu$, to model the function $f$. To keep things simple, we will use a notation that fits in our framework. We will denote the posterior by $q(\vu, f)$ and the prior by $p_0^\hyp(\vu, f)$ where $\hyp$ is the set of kernel hyperparameter. We will denote the negative likelihoods by $\loss_t(f)$ which are assumed to be Gaussian. With this notation, we assume that $q_t$ and $\delta_t$ is given and our goal is to update $q_{t+1}$ as well as $\delta_{t+1}$, but we assume the
inducing variables $\vu$ remain at the same locations.

Because the model is conjugate, the sites are not approximate, that is, $\losshat_{i|t} = \loss_i$. With this, we can write the following (denoting the normalizing constant by $\mathcal{Z}_t(\delta_t)$),
\[
   q_t = \frac{1}{\mathcal{Z}_t(\hyp_t)} p_0^{\hyp_t} \prod_{i=1}^t \exp(-\loss_t) 
\]
With the goal of updating $q_{t+1}$ can be expressed as a posterior correction,
\begin{equation}
   \begin{split}
      q_{t+1} &= \argmin_{q \in \setDist } \,\, \sum_{i=1}^{t+1} \myexpect_q[ \loss_i] + \dkls{}{q}{p_0}  + \log \mathcal{Z}_{t+1}(\hyp)\\
      &= \argmin_{q \in \setDist } \,\, \log \frac{\mathcal{Z}_{t+1}(\hyp)}{\mathcal{Z}_t(\hyp_t)} + \myexpect_q \sqr{ \log \rnd{ \frac{p_0^{\hyp_t}}{p_0^{\hyp}} \frac{q}{q_t} \frac{1}{e^{-\loss_{t+1}}} } }    
   \end{split}
\end{equation}
where in the first line we added the explicit normalizing constant to show dependency on $\hyp$. The last line is written to show equivalence to \citet[Eq. 5, $\second$ line]{bui2017streaming}. Modification to handle non-conjugate likelihood is straightforwardly obtained by adding back the corrections for all the past $\loss_i$. The procedure can also benefit from adding memory replay, similarly to previous sections. A recent extension by \cite{ChangVJSK23} explores such directions using the
dual representations and finds good improvements. Additionally, a procedure by \citet{adam2021dual} is also a special instance of posterior correction to speed up hyperparameter learning in GPs.

%

\section{Conclusion and Discussion}

The ability to quickly adapt is the holy grail of intelligence and we propose posterior correction as a mechanism to instill this ability in learning algorithms. The suggestion is that, when adapting, the algorithm should aim to correct its posterior distribution. 
The proposal shares this view with those of \citet{zellner1988optimal} who derived Bayes' rule as an `optimal' information processing rule. There, the information is neither created nor destroyed, simply preserved. Our proposal extends this principle to approximate posteriors obtained by using the Bayesian learning rule of \citet{khanRue23}. The goal now is to preserve the information and to maintain optimality within a class of candidate posteriors. The corrections ensure
exactly this.
The message fits well into the current trend of deep learning where exact posteriors are impossible to compute, but the computation of approximate posterior is still feasible. The new proposal makes it possible to apply the existing Bayesian principle to all such new cases.

The paper also presents methods to correct the posterior. Various approximations with Taylor's method seem useful when computation and memory are limited, but new primal-dual techniques are also attractive to eventually aim for a near-perfect correction.
But, two questions still remain: what should the algorithm remember and what new experiences should it seek? We do not provide immediate answers to these but do set the scene to search for them. Essentially, memory should be chosen to minimize the corrections needed in the future, while new experiences should be chosen to enable easy-enough corrections (or at least not too daunting). Both seem extremely difficult because the future is unknown, but an algorithm
still has the freedom to explore the brave new world by choosing its future and fixing its past. Posterior correction sets the stage for all such explorations for algorithms that could some day have a natural human-like ability to adapt.

\section*{Acknowledgement}
I thank my family for their support throughout this work which started in 2019 and continued through the pandemic. This work is supported by JST CREST Grant Number JP-MJCR2112. I benefited immensely from many insightful discussions with many past and current members of the ABI team at RIKEN AIP as well as the CREST-ANR Bayes-Duality Project.
I specifically want to thank my colleagues Dr. T.~M{\"o}llenhoff (RIKEN AIP), Dr. Siddharth Swaroop (Harvard University), Ang Ming Liang (University College London), Nico Daheim (TU Darmstadt) and Dharmesh Tailor (University of Amsterdam).
Many thanks to Prof. H{\aa}vard Rue (KAUST) for his mentorship and encouragement throughout. 

\appendix

\section{Examples of Dual Parameterization of the VB Posterior}
\label{app:dualparam}

\subsection{A One-Dimensional Logistic Regression Example}
\label{app:oneDex}

\begin{figure}[h]
   \center
   \includegraphics[height=1.4in]{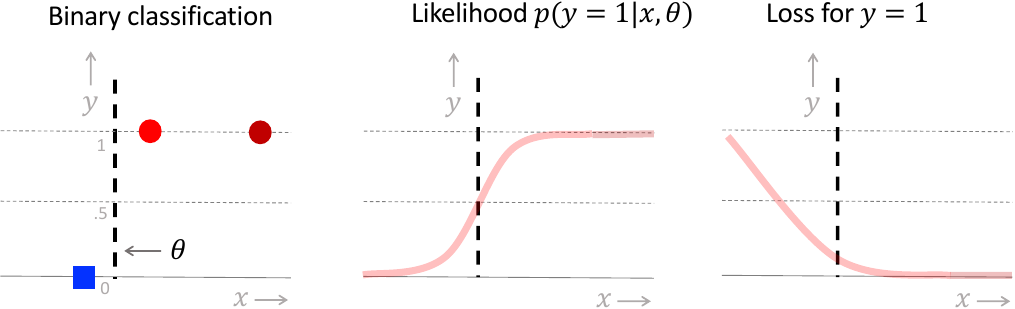}
   \caption{One dimensional binary classification, and its likelihood and loss function.}
   \label{fig:binary_classification}
\end{figure}

We give details on the illustration given in \cref{fig:dual_form} for a binary logistic-regression example. This is a 1-D example with a scalar input $x$ with a label $y\in\{0, 1\}$. We have three input-output pairs $(x_i, y_i)$, one for the square class (call it class 0) and two for the circle class (call it class 1). This is shown in the first panel of \cref{fig:binary_classification}. 
A classifier is simply a vertical line located at $\theta$. For instance, we can say that when $x>\theta$ then $y=1$, otherwise $y=0$. The logistic loss provides a `soft' thresholding via a sigmoid function,
\[
   \sigmoid(x-\theta) = \frac{1}{1+\exp(-(x-\theta))}.
\]
This is then used to define the likelihood function,
\[
   p(y_i|x_i, \theta) = \sigmoid(f_i)^{y_i} \sqr{1 - \sigmoid(f_i)}^{1-y_i},
\]
where $f_i = x_i - \theta$ are example dependent `regression' functions, also known as \emph{logits}. The loss function is simply the negative-log of the likelihood,
\[
   \loss_i(\theta) = -\log p(y_i|x_i, \theta) = -y_i \log \sigmoid(f_i) - (1-y_i) \sqr{1 - \sigmoid(f_i)}.
\]
This is the two class version of the well-known cross-entropy loss.
The likelihood and loss functions are visualized in the second and third panel of \cref{fig:binary_classification} for $y=1$ as a function of $x$.

The goal of learning is to find $\theta$ such that all examples are classified well, which can be formulated as a (regularizerd) empirical risk minimization,
\[
   \theta_t = \arg\min_{\theta} \sum_{i=1}^t \loss_i  + \half \theta^2.
\]
The first panel in \cref{fig:losses} below illustrates all the losses $\loss_i$ while the second panel shows the ERM. Note that the $x$-axis here represents $\theta$ in contrast to \cref{fig:binary_classification} where we plotted the likelihood functions against $x$. The second panel shows the overall loss and minimizer which is located between $x_1$ and $x_2$.
\begin{figure}[h]
   \center
   \includegraphics[width=6in]{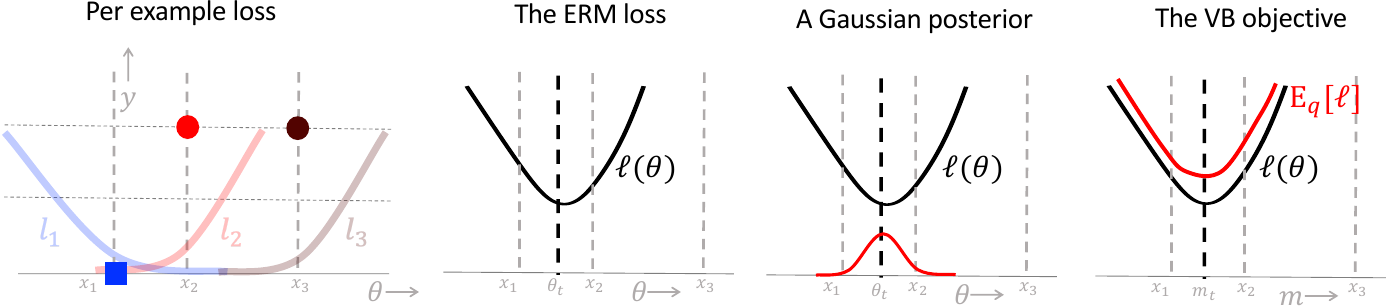}
   \caption{The first panel shows the losses $\loss_i$ for the three examples. The second panel shows the overall ERM and the minimizer $\theta_t$. The third panel shows a Gaussian posterior $q(\theta)$ with mean $m = \theta_t$. The fourth panel shows the expectation of the ERM over $q_t$.}
   \label{fig:losses}
\end{figure}

The third panel illustrates a Bayesian approach where the goal is to locate a Gaussian posterior $q(\theta) = \gauss(\theta|m, \sigma^2)$ whose mean is close to $\theta_t$. The resulting loss function uses an expected loss $\myexpect_q[\loss]$ which can be plotted as a function of the mean $m$. This is shown in the fourth panel where we see that the expected loss closely follows the ERM loss. The higher values are due to averaging over the neighborhood of $m$. The minimizer
$m_t$ is close to $\theta_t$ which can be recovered from the VB objective by essentially ignoring the expectation (via the delta method as explained in the main text).

We will now derive the form of the site functions and show that the Gaussian posterior can be written as a product of the three sites. We denote the $L_2$ regularizer $\loss_0 = \half \theta^2$. As shown in \cref{tab:site}, the site function for a Gaussian $q_t = \gauss(m_t, \sigma_t^2)$ is 
\[
   \losshat_{i|t} = \theta \cdot \myexpect_{q_t}[ \nabla \loss_i] + \half (\theta-m_t)^2 \cdot \myexpect_{q_t}[ \nabla^2 \loss_i] + \cnst
\]
The expectations of the gradient and Hessian of the logistic loss can be simplified by noting the fact that $\sigmoid'(f) = \sigmoid(f) [1-\sigmoid(f)]$, as shown below
\begin{equation}
   \begin{split}
      \myexpect_{q_t}[\nabla \loss_i(\theta)] &=  \myexpect_{q_t} \sqr{ \rnd{ \frac{y_i}{\sigmoid(f_i)} - \frac{1-y_i}{1-\sigmoid(f_i)} } \sigmoid'(f_i) } 
      = \sqr{ y_i - \myexpect_{q_t}\rnd{\sigmoid(f_i)} } := \alpha_{i|t}\\ 
      \myexpect_{q_t}[\nabla^2 \loss_i(\theta)] &= -\myexpect_{q_t} \sqr{ \nabla\sigmoid(f_i) } = \myexpect_{q_t} \sqr{ \sigmoid'(f_i) } := \beta_{i|t}
   \end{split}
\end{equation}
For simplicity, we denote the two quantities by $\alpha_{i|t}$ and $\beta_{i|t}$. We note that $\alpha_{i|t}\in (-1,1)$ is also the residual and $\beta_{i|t} \in (0, 1/4)$, a strictly positive scalar.

\begin{figure}[h]
   \center
   \includegraphics[width=4.5in]{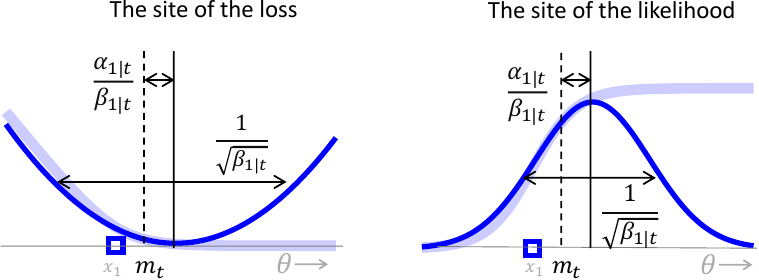}
   \caption{The site functions (in dark blue) for the loss (left) and likelihood (right), both shown in light blue.}
   \label{fig:site1}
\end{figure}

The site for the loss and likelihood can respectively be written as  
\begin{align*}
   &\losshat_{i|t} = \theta \alpha_{i|t} + \half \beta_{i|t} (\theta-m_t)^2 + \cnst =  \half \beta_{i|t} \rnd{ \theta - \ty_{i|t} }^2 + \cnst\\
   &\implies
   \exp(-\losshat_{i|t}) \propto \gauss(\theta| \ty_{i|t},  1/\beta_{i|t})
\end{align*}
where $\ty_{i|t} = m_t - \alpha_{i|t}/\beta_{i|t} $. \cref{fig:site1} illustrates the site function of $x_1$ which is located on the left side of $m_t$. 
The pair $(\alpha_{i|t}, \beta_{i|t})$ determine the location and curvature, respectively. Essentially, the location of the quadratic shown on the left is $|\frac{\alpha_{i|t}}{\beta_{i|t}}|$ distance away from the mean $m_t$. Whether it is on the left or right depends on the sign of the ratio. The variance
of the Gaussian factor is equal to $1/\beta_{i|t}$. The quantity $\ty_{i|t}$ can be seen as the location or as a pseudo-Gaussian observation. Note that the Gaussian factor over $\theta$ is not always guaranteed to be a valid probability `distribution' over $\theta$.

\begin{figure}[h]
   \center
   \includegraphics[width=4in]{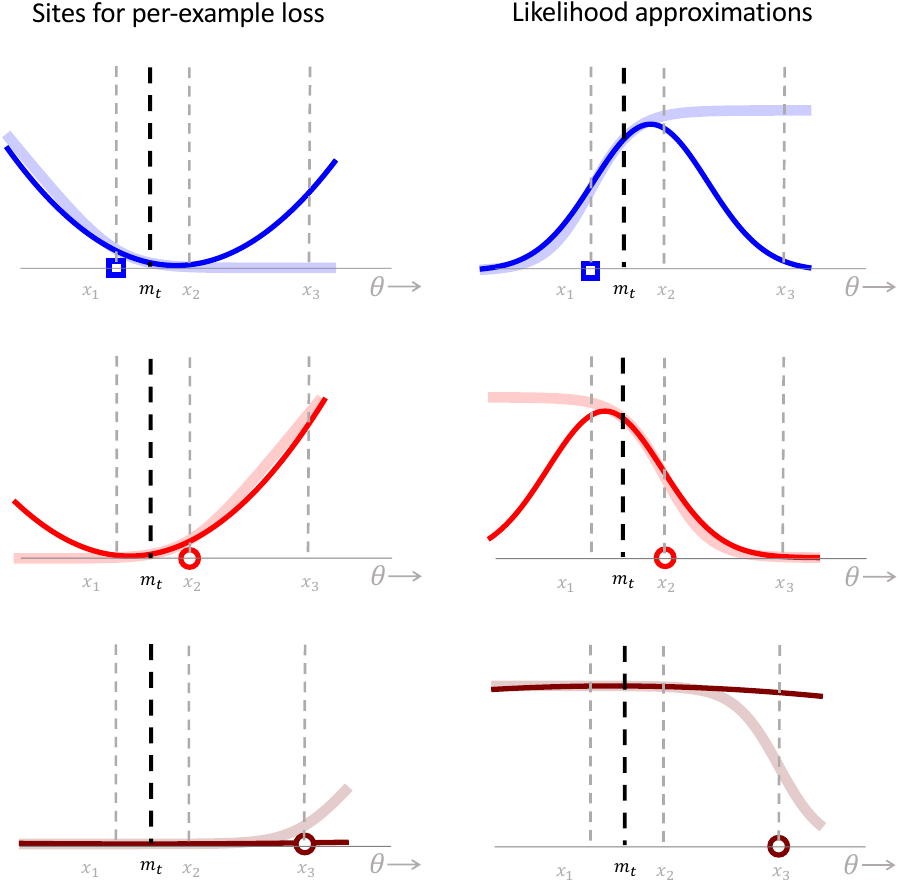}
   \caption{Each row shows the sites of an example. Sites for the loss are shown on the left while those for the likelihood are on the right.}
   \label{fig:sites_ex}
\end{figure}

\cref{fig:sites_ex} compares the sites for the losses and likelihoods of all the three example. The site of the observations $x_1$ and $x_2$ both which are close to $m_t$ have a sharper curvature than $x_3$ which is far away. The location of the observation also decides the location of the sites. The input $x_1$ is on the left side of $m_t$ and its site is located on the other side, while the opposite is true for $x_2$. The site for $x_3$ is located much farther
from $m_t$.

From the figure above it is clear that the sites of $x_1$ and $x_2$ are much more prominent than that of $x_3$. As a result, the contributions of each site to $q_t$ also differs drastically from each other. We will now present precise expressions that show that the example $(x_3, y_3)$ contributes a lot less to $q_t$ than the other two examples. In fact, the decision would change very little even if $(x_3,y_3)$ were to be removed.
To see this, let us write the dual representation of the posterior in terms of the site functions,
\begin{align}
   \gauss(\theta|m_t, \sigma_t^2) &= \frac{1}{\mathcal{Z}_t} \exp\rnd{-\half \theta^2} \prod_{i=1}^t \exp\sqr{ - \half \beta_{i|t} \theta^2 + \ty_{i|t} \beta_{i|t} \theta } 
   \label{eq:qdualform_1d}
\end{align}
This is illustrated in the figure below (where we ignored the prior term) to show how Gaussian factors are multiplied together to find the posterior over $\theta$.
\begin{figure}[h]
   \center
   \includegraphics[width=4in]{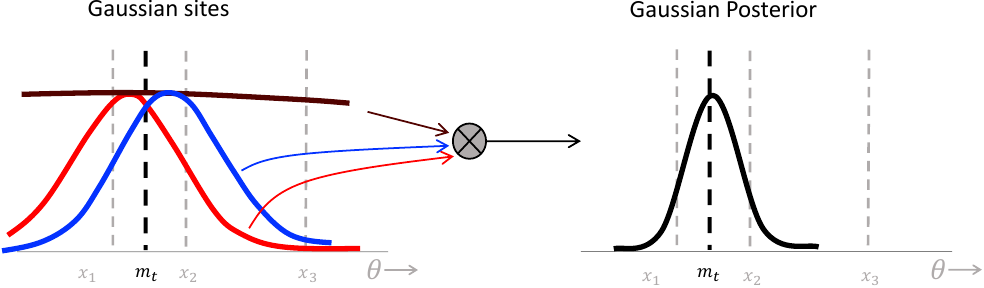}
   \caption{Multiplication of the sites followed by normalization yields the posterior.}
   \label{fig:final_dist}
\end{figure}

We will express the mean $m_t$ and variance $\sigma_t^2$ in terms of the site parameters $(\alpha_{i|t}, \beta_{i|t})$. Both sides in \cref{eq:qdualform_1d} have Gaussian forms, therefore we can expand them to see how each of them contributes to the sufficient statisics $(\theta, \theta^2)$ of the posterior, 
\begin{align*}
   \exp\sqr{ -\half \rnd{ \frac{1}{\sigma_t^{2} } } \theta^2 + \rnd{ \frac{m_t}{\sigma_t^{2}} } \theta} &= \frac{1}{\mathcal{Z}_t} \exp\sqr{ - \half \theta^2} \prod_{i=1}^t \exp\sqr{ - \half \beta_{i|t} \theta^2 + \ty_{i|t} \beta_{i|t} \theta } \\ 
   &= \frac{1}{\mathcal{Z}_t} \exp\sqr{  -\half \rnd{ 1+ \sum_{i=1}^t \beta_{i|t} } \theta^2 + \rnd{ \sum_{i=1}^t \ty_{i|t} \beta_{i|t} } \theta } \\
   &= \frac{1}{\mathcal{Z}_t} \exp\sqr{  -\half \rnd{ 1+ \sum_{i=1}^t \beta_{i|t} } \theta^2 + \rnd{m_t \sum_{i=1}^t \beta_{i|t} -  \sum_{i=1}^t \alpha_{i|t} } \theta }. 
\end{align*}
Matching the terms in front of $\theta^2$, we find that the posterior precision $1/\sigma_t^2$ is expressed as the sum of individual precisions $\beta_{i|t}$, that is,
\begin{equation}
   \frac{1}{\sigma_t^2} = 1 + \sum_{i=1}^t \beta_{i|t}. 
   \label{eq:repr_var_1d}
\end{equation}
Then matching the terms in front of $\theta$ and using the last expression, we get
\begin{align}
   \frac{m_t}{\sigma_t^2} &= m_t \sum_{i=1}^t \beta_{i|t} -  \sum_{i=1}^t \alpha_{i|t} = m_t \rnd{\frac{1}{\sigma_t^2} - 1} -  \sum_{i=1}^t \alpha_{i|t}
   \quad&\implies \quad 
   m_t = -\sum_{i=1}^t \alpha_{i|t}
\end{align}
Readers familiar with the representer theorem may see the above formula as an instance of the well-known representer theorem from Kernel methods \citep{scholkopf2001generalized}. This is now recovered by matching the sufficient statistics of the posterior with those of the dual form. The dual form presented here is more general because it also yields a representation formula for the variance as shown in \cref{eq:repr_var_1d}.

\subsection{Binary Logistic Regression: General Case}
\label{app:logreg}

Consider a binary logistic regression problem with $N$ input-output pairs $(\vx_i, y_i)$. For this case, we can define the logits as linear predictors $f_i = \vx_i^\top\vparam$. We will derive the form of the site functions for a full Gaussian distribution $q_t = \gauss(\vparam|\vm_t, \vSigma_t)$. The site function are defined through $\myexpect_{q_t}[\nabla \loss_i]$ and $\myexpect_{q_t}[\nabla^2 \loss_i]$, both of which can be simplified. For example, we can change the variable from $\vparam$ to $f$ to write
\begin{equation*}
   \myexpect_{q_t}[\nabla \loss_i] = \int \underbrace{ \vx_i \loss_i'(\vx_i^\top\vparam) }_{=\nabla\loss_i(\vparam)} \underbrace{ \gauss(\vparam|\vm_t, \vSigma_t) }_{=q_t(\vparam)} d\vparam 
   = \vx_i  \int \loss_i'(f) \underbrace{ \gauss(f|\vx_i^\top\vm_t, \vx_i^\top\vSigma_t \vx_i) }_{= \tq_{i|t}(f)} df 
   = \vx_i \bar\alpha_{i|t}. 
\end{equation*}
Here, $\tq_{i|t}(f)$ is a distribution over $f=\vx_i^\top\vparam$ derived from $q_t(\vparam)$ and we define 
\[
   \bar\alpha_{i|t} = \myexpect_{\tq_{i|t}} [\loss_i'] \text{ where } \loss_i'(f_i) = p(f_i) - y_i 
\]
is the residual. In a similar way, we can write \smash{$\myexpect_{q_t}[\nabla^2 \loss_i] = \vx_i \vx_i^\top \bar\beta_{i|t}$}, where we define
\[
   \bar\beta_{i|t} = \myexpect_{\tq_{i|t}} [\loss_i''] \text{ with } \loss_i''(f_i) = p(f_i) [1-p(f_i)].
\]
Thus, we get write the site function as 1-dimension function over $f_i(\vparam) = \vx_i^\top\vparam$,
\begin{equation}
   \begin{split}
      \losshatfull_{i|t}(\vparam) &= \vparam^\top \vx_i \bar\alpha_{i|t} + \half (\vparam-\vm_t)^\top \vx_i \bar\beta_{i|t} \vx_i^\top (\vparam-\vm_t)  \\
      &= \bar\alpha_{i|t} f_i(\vparam)  + \half \bar\beta_{i|t} (f_i(\vparam) - \tm_{i|t} )^2  + \cnst \\
      &= \half \beta_{i|t} (f_i(\vparam) - \ty_{i|t})^2  + \cnst
   \end{split}
\end{equation}
Here, we defined $\tm_{i|t} = \vx_i^\top \vm_t$ and $\ty_{i|t} = \tm_{i|t} - \bar\alpha_{i|t}/ \bar\beta_{i|t}$. Similarly to the simple 1-dimensional case, the site is a quadratic function whose location is $|\bar\alpha_{i|t} / \bar\beta_{i|t}|$ distance away from $\tm_{i|t}$ and its curvature is $\bar\beta_{i|t}$. The site can also be written as a Gaussian factor,
\[
   \exp(-\losshat_{i|t}) \propto \gauss(\theta| \ty_{i|t},  1/\bar\beta_{i|t})
\]
With this site function, we can write the dual form of the posterior by using \cref{eq:qdualform} (and assuming a standard normal prior),
\begin{align*}
   \gauss(\vparam|\vm_t, \vSigma_t) &= \frac{1}{\mathcal{Z}_t} \exp\rnd{-\half \vparam^\top\vparam} \prod_{i=1}^t \exp\sqr{ - \half \bar\beta_{i|t} f_i(\vparam)^2 + \ty_{i|t} \bar\beta_{i|t} f_i(\vparam) } 
\end{align*}
We can also express the mean $\vm_t$ and covariance $\vSigma_t$ in terms of the site parameters $(\bar\alpha_{i|t}, \bar\beta_{i|t})$. There are two ways to go about it by using the two forms given in \cref{eq:deriv_dual}, one using the natural parameter form and the other using the distributional form. Here, we will use the latter and match terms in front of the sufficient statistics $\vparam$ and $\vparam\vparam^\top$, as shown below:
\begin{align*}
   \exp\sqr{ -\half \vparam^\top \vSigma_t^{-1} \vparam + \vm_t^\top\vSigma_t^{-1}\vparam} &= \frac{1}{\mathcal{Z}_t} \exp\sqr{ - \half \vparam^\top\vI\vparam} \prod_{i=1}^t \exp\sqr{ - \half \vparam^\top \vx_i \bar\beta_{i|t} \vx_i^\top \vparam + \ty_{i|t} \bar\beta_{i|t} \vx_i^\top\vparam } \\ 
   &= \frac{1}{\mathcal{Z}_t} \exp\sqr{  -\half \vparam^\top \rnd{ \vI + \sum_{i=1}^t \vx_i \bar\beta_{i|t} \vx_i^\top} \vparam + \rnd{ \sum_{i=1}^t \ty_{i|t} \bar\beta_{i|t} \vx_i } \vparam} 
\end{align*}
Matching the coefficients of $\vparam\vparam^\top$ gives us,
\begin{equation}
   \vSigma_t^{-1} = \vI + \sum_{i=1}^t \vx_i\bar\beta_{i|t}\vx_i^\top .
   \label{eq:Sigma_logreg_dual}
\end{equation}
Similarly, matching the coefficients of $\vparam$, we get
\begin{align*}
   \vSigma_t^{-1} \vm_t 
   = \sum_{i=1}^t (\bar\beta_{i|t} \tm_{i|t} - \bar\alpha_{i|t}) \vx_i 
   = \rnd{ \sum_{i=1}^t \vx_i \bar\beta_{i|t} \vx_i^\top } \vm_t - \sum_{i=1}^t \bar\alpha_{i|t} \vx_i 
   = \rnd{ \vSigma_t^{-1} - \vI} \vm_t - \sum_{i=1}^t \bar\alpha_{i|t} \vx_i 
\end{align*}
This gives us the following representer theorem,
\begin{equation}
   \vm_t = - \sum_{i=1}^t \bar\alpha_{i|t} \vx_i.
\end{equation}
This result is analogues to the well-known representer theorem, specialized to finite dimensional features. The form used there can be obtained by simply applying an arbitary test input $\vx_*^\top$ on both sides, and defining $f_{*|t} = \vx_*^\top\vm_t$ and $\kappa(\vx_*, \vx_i) = \vx_*^\top\vx_i$. This gives us
\begin{equation}
   f_{*|t} = - \sum_{i=1}^t \bar\alpha_{i|t} \kappa(\vx_*, \vx_i).
\end{equation}
By using delta method, we can approximate $\bar\alpha_{i|t} \approx \loss'(\tm_{i|t})$ which recovers exactly the result of the representer theorem. Here too, our dual form is more general because we also get a representation formula for the covariance, as shown in \cref{eq:Sigma_logreg_dual}.

\subsection{Dual Forms During Training for Logistic Regression}
\label{app:dual_training}

The dual form given in \cref{eq:qdualform} also can be written during the iteration of the BLR as shown in \cref{eq:deriv_dual}. We will now give an example for the logistic regression case.
\[
   q_t \leftarrow \frac{1}{\mathcal{Z}_t} q_t^{1-\rho}\prod_{i=0}^t \exp\rnd{- \rho\losshat_{i|t} } 
\]
As shown in \cite[Sec. 3.1.2]{khanRue23}, the update can be written in a Newton-like form, by first defining the precondition matrix via the covariance,
\[
   \rnd{\vSigma_t^\new}^{-1} \gets (1-\rho)\vSigma_t^{-1} + \rho \rnd{ \vI + \sum_{i=1}^t \vx_i\bar\beta_{i|t}\vx_i^\top },
\]
and then using it to update of the mean,
\begin{equation}
   \vm_t \gets \vm_t - \rho \rnd{\vSigma_t^{\new}}^{-1} \rnd{ \sum_{i=1}^t \bar\alpha_{i|t} \vx_i } 
\end{equation}
Multiplying both sides by $\vx_*^\top$, we can write the update in the function space in terms of individual contributions from each data example, 
\[
   f_{*|t} \gets f_{*|t} - \rho \sum_{i=1}^t \bar\alpha_{i|t} \kappa(\vx_*, \vx_i) 
   \qquad \text{ where } \quad
   \kappa(\vx_*, \vx_i) = \vx_*^\top\rnd{ \vSigma_t^\new}^{-1} \vx_i.
\]
A derivation for neural network is in \citet{khan2019approximate} by using Gauss-Newton approximation, and a slightly more general form is given in \citet[App. C.3]{nickl2023memory}.

\section{Extension of the Dual Form to EF Distributions with Non-Constant Base Measures}
\label{app:base_measure}

For distributions with non-constant base measures $h(\vparam)$, only a small change in the second step is required. Consider the family $q(\vparam) = h(\vparam) \exp(\myang{\vT(\vparam), \vlambda} - A(\vlambda))$ with the base
measure, then the natural gradient of the entropy term \citep[App. B]{khanRue23} also contains an additional term $\natgrad \myexpect_{q_t}[\log h]$.
This can be added in the existing dual form by simply redefining the regularizer site as follows:
\begin{equation}
   \losshat_{0|t}(\vparam)  = \myang{\vT(\vparam) , \natgrad \myexpect_{q_t}[\loss_0 + \log h ]} - \log h(\vparam).
   \label{eq:def_l0hat}
\end{equation}

\section{Expected Hessian for a Bregman Loss}
\label{app:qfull_corr_cl}

For Bregman losses, we know that the Hessian involves a Generalized Gauss-Newton (GGN) matrix and the Hessian of the model output \citep[Eq. 6]{martens2020}. Using this, we can then apply delta method to simplify as follows,
\begin{equation}
   \begin{split}
      \bar\vH_{i|t} &= \myexpect_{q_t} \sqr{ \nabla^2 \glmloss [ y_i, \hat y_i(\vparam) ] } \\
      &= \myexpect_{q_t} \sqr{  \nabla f_i(\vparam) \,\, {\left. \nabla^2 \glmloss[y_i, z] \right\vert_{z = f_i(\vparam)} } \,\, \nabla f_i(\vparam)^\top +   {\left. \nabla \glmloss[y_i, z] \right\vert_{z = f_i(\vparam)} } \,\, \nabla^2 f_i(\vparam) } \\
      &\approx \nabla f_i(\vm_t) \beta_{i|t} \nabla f_i(\vm_t)^\top +   r_{i|t} \nabla^2 f_i(\vm_t).
   \end{split}
\end{equation}
where we use the fact that the first derivative of the Bregman loss with respect to $z$ at $z = f_i(\vparam)$ is simply the residual of the prediction $\hat y_i(\vparam)$,
\begin{equation}
   \left. \nabla \glmloss[y_i, z] \right\vert_{z = f_i(\vparam)} = A'(f_i(\vparam)) - y_i = \hat y_i(\vparam) - y_i
   \quad \implies\,\,
   \left. \nabla^2 \glmloss[y_i, z] \right\vert_{z = f_i(\vparam)} = A''(f_i(\vparam)) ,
\end{equation}
where we denote $r_{i|t} = \myexpect_{q_t} [\hat y_i(\vparam) - y_i] $ and $\beta_{i|t} = \myexpect_{q_t} [A''(f_i(\vparam))]$.

\section{Derivation of the Correction Term of \cref{eq:inf_corr_l0} for $\loss_0$}
\label{app:inf_corr_l0}

The correction term for $\loss_0$ in the case of $\qiso$ takes a bit more effort because the base measure is not constant and is given by 
\[
   h(\vparam) = -\half \|\vparam\|^2 - \half P\log (2\pi).
\]
For such cases, the definition of the correction for $\loss_0$ is given in \cref{eq:def_l0hat}.
\begin{equation}
   \begin{split}
      \loss_0(\vparam) - \losshatiso_{0|t}(\vparam) &= \loss_0(\vparam) - \vparam^\top \myexpect_{q_t}[ \nabla \loss_0 + \nabla \log h] + \log h(\vparam) \\
      &= \loss_0(\vparam) - \vparam^\top \myexpect_{q_t}[ \nabla \loss_0] + \vparam^\top\vm_t - \half \|\vparam\|^2 \\
      &= \loss_0(\vparam) - \vparam^\top \myexpect_{q_t}[ \nabla \loss_0] - \half \|\vparam - \vm_t\|^2 + \cnst \\
      &= \loss_0(\vparam) - \vparam^\top \myexpect_{q_t}[ \nabla \loss_0]  - \dkls{}{q}{q_t} + \cnst \\
   \end{split}
   \label{eq:kprior_deriv_l0}
\end{equation}

\section{Derivation of \cref{eq:if_qfull} for Influence Estimation}
\label{app:if_qfull}

We start by writing posterior correction in \cref{eq:postcorr_influence} over $\qfull$ family in the first line and then ignore the correction term in the second line, followed by plugging in the definition of $\losshat_{j|t}$ and KL from \cref{tab:site} and \cref{eq:kl} respectively in the third line, and then simplify,
\begin{equation}
   \begin{split}
      &q_{t\remove j} = \argmin_{q \in \setDist^{\text{full}} } \,\, \myexpect_q[-\losshat_{j|t}] + \dkls{}{q}{q_t} + \sum_{i=0:t, i\ne j} \myexpect_q[\loss_i - \losshat_{i|t}] \\
      &\approx \argmin_{q \in \setDist^{\text{full}} } \,\, \myexpect_q[-\losshat_{j|t}] + \dkls{}{q}{q_t} \\
      &= \argmin_{\vm,\text{\vS} } \,\, \myexpect_q\sqr{ \vparam^\top \myexpect_{q_t}[-\nabla \loss_i] + \half (\vparam-\vm_t)^\top \myexpect_{q_t}[-\nabla^2 \loss_i] (\vparam-\vm_t)]  } \\
      &\quad\quad\quad\quad\quad\quad\quad\quad  + \half (\vm-\vm_t)^\top\vS_t(\vm-\vm_t) + \half \trace\sqr{\vS^{-1}\vS_t} - \half \log\left|\vS^{-1}\vS_t\right| \\
      &= \argmin_{\vm,\text{\vS} } \,\, \vm^\top \myexpect_{q_t}[-\nabla \loss_i] + \half (\vm-\vm_t)^\top \sqr{ \vS_t - \myexpect_{q_t}[\nabla^2 \loss_i] } (\vm-\vm_t)  \\
      &\quad\quad\quad\quad\quad\quad\quad\quad  + \half \trace\sqr{\vS^{-1} \rnd{ \vS_t - \myexpect_{q_t}[\nabla^2 \loss_i] } } - \half \log\left|\vS^{-1}\vS_t\right| \\
      &= \argmin_{\vm,\text{\vS} } \,\, \vm^\top \myexpect_{q_t}[-\nabla \loss_i] + \half (\vm-\vm_t)^\top \hat\vS_{t\remove j} (\vm-\vm_t) + \half \trace\sqr{\vS^{-1} \hat\vS_{t\remove j}  } - \half \log\left|\vS^{-1}\right| 
   \end{split}
\end{equation}
where $\hat\vS_{t\remove j} = \vS_t - \myexpect_{q_t}[\nabla^2 \loss_i] $. Then taking derivative over $\vm$ and $\vS^{-1}$, we get the desired result shown in \cref{eq:if_qfull}.

\section{Derivation of \cref{eq:ha} for Model Merging}
\label{app:deriv_merging}

We give a detailed derivation of \cref{eq:ha}. We first derive expressions for the correction terms by using the $\second$-order Taylor expansion and apply the delta method to cancel the first term with the last term,
\[
   \loss_i - \losshat_{i} \approx  \vparam^\top \nabla \loss_i(\vm_i) + \half (\vparam - \vm_i)^\top \vH_i (\vparam - \vm_i) - \vparam^\top \myexpect_{q_i}[\nabla \loss_i] \approx \underbrace{ \half (\vparam - \vm_i)^\top\vH_i (\vparam - \vm_i) }_{:= -\log \hat{q}_i(\vparam)}
\]
We denote the terms by the log of a Gaussian factor $\hat{q}_i$. Approximation for the correction of $\loss_\llm$ is obtained in a similar way, and we will denote it by $\hat{q}_\llm$. Plugging these in \cref{eq:postcorr_merging}, we can rewrite the posterior merging as 
\begin{align*}
      q_{\text{retrained}} & =\argmin_{q \in \mathcal{Q} } \,\, \dkls{}{q}{q_\ba} + \sum_{i=1}^t \alpha_i \myexpect_{q}[\loss_i - \losshat_{i}]  + \myexpect_q[\loss_{\llm} - \losshat_{\llm}] \\
      &\approx \argmin_{q \in \mathcal{Q} } \,\, \dkls{}{q}{q_\ba} - \sum_{i=1}^t \alpha_i\myexpect_q[ \log \hat{q}_i] - \myexpect_q[\log \hat{q}_\llm ] \\
      &= \argmin_{q \in \mathcal{Q} } \,\, \mathbb{D}_{\text{KL}} \sqr{ q \| \frac{q_\llm \hat{q}_\llm}{Z_t} \prod_{i=1}^t \rnd{ \frac{q_i \hat{q}_i}{q_\llm}}^{\alpha_i} }
\end{align*}
   where $Z_t$ is the normalizing constant.
   The solution is when $q$ is equal to the distribution at in the right argument of the KL term. Then, using the formula for Gaussian multiplication and division, 
   \begin{align*}
      &\gauss(\vparam|\vm_{1+2-3}, \vH_{1+2-3}^{-1}) = \frac{\gauss(\vparam|\vm_1, \vH_1^{-1})\gauss(\vparam|\vm_2, \vH_2)}{\gauss(\vparam|\vm_3, \vH_3)} \\
      &\implies
      \vm_{1+2-3} = \vH_{1+2-3}^{-1} \rnd{ \vH_1\vm_1 + \vH_2 \vm_2 - \vH_3\vm_3} \\
      &\quad\quad\quad \vH_{1+2-3} = \vH_1 + \vH_2 - \vH_3,
   \end{align*}
we recover \cref{eq:ha} as shown below:
\begin{align*}
   \vH_\ha &= (\vI + \vH_\llm) + \sum_{i=1}^t \alpha_i \rnd{ \vI + \vH_i - \vI} = \vI + \vH_\llm + \sum_i \alpha_i \vH_i \\
   \vH_\ha\vm_\ha &= \vm_\llm + \vH_\llm\vm_\llm + \sum_{i=1}^t \alpha_i (\vm_i + \vH_i\vm_i - \vm_\llm) \\
   &= \vH_\ha \vm_\llm + \sum_{i=1}^t \alpha_i (\vI + \vH_i) (\vm_i - \vm_\llm)
\end{align*}

\section{Proofs for Federated Learning}

\subsection{Proof of Optimality for the Two-Step Posterior Correction Algorithm}
\label{app:fl_details}

If the algorithm converges to a pair $(\hat q_\jnt^*, q_i^*)$, then the sites also converge, say, to 
\[
   \losshat_{i}^* = \exp\rnd{- \myang{\vT(\vparam), \dualparamhat_{i}^* }},
   \quad \text{ where }\quad
   \dualparamhat_{i}^* = \natgrad \myexpect_{q_i^*}[\loss_i].
\]
Then, it immediately follows from the optimality condition of \cref{eq:local_update} that
\[ 
   q_i^* \propto \hat q_\jnt^* \exp\sqr{ - \frac{1}{\rho} ( \loss_{i}^* - \loss_{i}^* ) } = \hat q_\jnt^*, \quad \forall i. 
\]
Essentially, the corrections vanish. Due to this equality, \cref{eq:global_update} yields
\[
   \hat{q}_\jnt^* \gets \frac{1}{\hat{\mathcal{Z}}_\jnt^*} p_0 \prod_{i=1}^t \exp\rnd{- \myang{\vT(\vparam), \natgrad \myexpect_{\hat q_\jnt^*}[\loss_i]} }. 
\]
showing that $\hat{q}_\jnt^*$ satisfies the optimality condition in \cref{eq:fl_joint}, thus completing the proof.

\subsection{Alternating Minimization Algorithm (AMA)}
\label{app:ama}

We will first derive the update shown in \cref{eq:ama_update} and then show similarity to posterior correction. In AMA, we reformulate the problem as the following constrained reformulation by splitting,
\[
   \argmin_{\vparam_\jnt, \vparam_1, \vparam_2, \ldots, \vparam_t} \sum_{i=1}^t \loss_i(\vparam_i ) + \half \| \vparam_\jnt \|^2, 
   \quad \text{ such that }  \vparam_i = \vparam_\jnt, \quad \forall i=1,2,\ldots, t.
\]
A Lagrangian is formed for the above problem using dual vector $\vv_i$ for each constraints, which gives rise to the following three steps to update the triplet $(\vparam_\jnt, \vparam_i, \vv_i)$,
\[
   \mathcal{L}_{\text{lag}}(\vparam_\jnt, \vparam_{1:t}, \vv_{1:t}) = \sum_{i=1}^t \loss_i( \vparam_i ) + \half \|\vparam_\jnt\|^2 + \sum_{i=1}^t \vv_i^\top(\vparam_\jnt - \vparam_i),
\]
where we denote the vectors of $\vparam_i$ and $\vv_i$ by $\vparam_{1:t}$ and $\vv_{1:t}$ respectively. To update $\vparam_\jnt$, we fix $\vv_i$ and $\vparam_{1:t}$ and set the derivative with respect to $\vparam_\jnt$ to zero to get
\[
   \vparam_\jnt \leftarrow \argmin_{\vparam_\jnt} \half \|\vparam_\jnt\|^2 + \sum_{i=1}^t \vv_i^\top(\vparam_\jnt - \vparam_i),
   \quad\implies\quad
   \vparam_\jnt \gets \sum_{i=1}^t \vv_i.
\]
For $\vparam_i$, AMA uses an `augmented' Lagrangian, where we collect all the terms in the Lagrangian that depend on $\vparam_i$ and add a quadratic proximal term with a coefficient $\rho>0$,
\begin{equation}
   \begin{split}
      \vparam_i &\gets \argmin_{\vparam_i} \,\, \loss_i(\vparam_i) + \vv_i^\top(\vparam_\jnt - \vparam_i) + \rho \half \|\vparam_\jnt- \vparam_i \|^2 \\
      &= \argmin_{\vparam_i} \,\, \loss_i(\vparam_i) - \vv_i^\top \vparam_i + \rho \half \|\vparam_\jnt- \vparam_i \|^2 .
   \end{split}
   \label{eq:ama_thetai_update}
\end{equation}
This gives us the update of $\vparam_i$.

Finally, the update of the dual vector $\vv_i$ is done as follows,
\begin{equation}
   \vv_i \gets \vv_i^\old + \rho(\vparam_i - \hat\vparam_\jnt) ,
\end{equation}
which can be simplified by noting the optimality condition of \cref{eq:ama_thetai_update}, where we have the following if we denote the old $\vv_i$ by $\vv_i^\old$,
\[
   \nabla \loss_i(\vparam_i) - \vv_i^\old + \rho (\hat\vparam_\jnt - \vparam_i)
   \quad\implies\quad
   \nabla \loss_i(\vparam_i) = \vv_i^\old + \rho (\hat\vparam_\jnt - \vparam_i).
\]
From here it follows that $\vv_i \gets \nabla \loss_i(\vparam_i)$.

\subsection{AMA as Posterior Correction}
\label{app:amaaspc}

We will first write double-loop posterior correction in the form shown in \cref{eq:2stagePoCoNatparamForm}.
\begin{equation}
   \begin{split}
      q_\jnt &\gets \frac{1}{\hat{\mathcal{Z}}_\jnt} p_0 \prod_{i=1}^t \exp\rnd{-\losshat_{i}} \\
      &= \frac{1}{\hat{\mathcal{Z}}_\jnt} \exp\rnd{\myang{\vT(\vparam), \vnatparam_0} } \prod_{i=1}^t \exp\rnd{-\myang{\vT(\vparam), \dualparamhat_{i}}} \\
      &= \frac{1}{\hat{\mathcal{Z}}_\jnt} \exp\rnd{-\myang{\vT(\vparam), \sum_{i=0}^t \dualparamhat_{i}}},
   \end{split}
\end{equation}
where in the second line we used the fact that the prior $p_0$ takes the same EF form as the $q_\jnt$ with natural parameter $\vnatparam_0$. In the third line, we denote $\dualparamhat_{0|0} = \vnatparam_0$. From here it follows that the natural parameter of $q_\jnt$ is set to $\vnatparam_\jnt \gets \sum_{i=0}^t \dualparamhat_i$.

Next, we rewrite \cref{eq:local_update} in terms of the expectation parameters $\vmeanparam_i$ of $q_i$,
\begin{equation}
   \begin{split}
      &q_i \gets \argmin_{q_i \in \mathcal{Q} } \,\, \frac{1}{\rho} \myexpect_{q_i} \sqr{\loss_i - \myang{\vT(\vparam), \dualparamhat_{i}} } +  \dkls{}{q_i}{q_\jnt} \\
      &\implies \vmeanparam_i = \argmin_{\vmeanparam_i } \,\,\myexpect_{q_i} \sqr{\loss_i} - \myang{\vmeanparam_i, \dualparamhat_{i}} +  \rho \dkls{}{q_i}{q_\jnt}. 
   \end{split}
\end{equation}
The equivalence in second line is due to the fact that, similarly to $\vnatparam_i$, $\vmeanparam_i$ is a valid parameterization of $q_i$ which follos from the fact that the mapping between the pair $(\vnatparam_i, \vmeanparam_i)$ is a bijection.

We will now show that AMA is recovered as a special case if we use the $\qiso$ family for posterior correction.
We start with the update of $\vparam_\jnt$ which can be recovered from the outer loop of the posterior correction algorithm. We define $q_\jnt = \gauss(\vparam|\vm_\jnt, \vI)$ with the mean $\vm_\jnt$. Noting that for $\qiso$ we have $\vT(\vparam) = \vparam$, we can write the sites as $\losshat_{i}(\vparam) = \vparam^\top \dualparamhat_{i}$. As a result, the update in \cref{eq:global_update} becomes
\[
   \gauss(\vparam|\vm_\jnt, \vI) \propto \gauss(\vparam|0, \vI) \prod_{i=1}^t \exp\rnd{- \vparam^\top \dualparamhat_{i}} 
   \quad \implies \quad
   \vm_\jnt \gets \sum_{i=1}^t \dualparamhat_{i}.
\]
Clearly, the site parameter $\dualparamhat_{i}$ serves the same role as the dual vector $\vv_i$.

Next, we recover the update of $\vparam_i$ which can be written as the inner loop of the posterior correction algorithm. We denote $q_i = \gauss(\vparam|\vm_i, \vI)$, then using the definition of the site and KL divergence, \cref{eq:local_update} reduces to
\begin{equation}
   \vm_i = \argmin_{\vm_i} \,\, \myexpect_{\text{\gauss}(\text{\vepsilon}|0, \text{\vI})}[\loss_i(\vm + \vepsilon)] - \dualparamhat_i^\top \vm_i + \rho \half\|\vm_\jnt- \vm_i \|^2 .
\end{equation}
If we set $\vepsilon = 0$, then the above equation coincides with the AMA update (after renaming $\vm_i$ and $\vm_\jnt$ by $\vparam_i$ and $\vparam_\jnt$ respectively. Finally, the update of the dual vector is also the same if we approximate the natural gradient as  $\myexpect_{q_i}[\nabla \loss_i] \approx \nabla \loss(\vm_i)$. This proves the result.

\subsection{PVI as Federated Posterior Correction}
\label{app:pvi_deriv}

We start with the second line of PVI in \cref{eq:pvi}, for which we need to write the optimality condition of the first line. We note that $\site_i \propto \exp(-\myang{\vT(\vparam), \dualparamhat_{i})})$, and denote the natural parameters of $q_i$ and $q_\jnt$ by $\vnatparam_i$ and $\vnatparam_\jnt$. We will also call $s_i$ as $s_i^\old$.
With this notation, the optimality of the first equation is obtained by taking natural gradients,
\begin{equation}
   \begin{split}
      &\natgrad \myexpect_{q_i} \sqr{\loss_i + \log \site_i^\old} + \natgrad  \dkls{}{q_i}{q_\jnt}  =0 \\
      \implies &\natgrad \myexpect_{q_i} \sqr{\loss_i} - \dualparamhat_{i} + \vnatparam_i - \vnatparam_\jnt  =0\\
      \implies &\vnatparam_i = \vnatparam_\jnt + \dualparamhat_{i} - \natgrad \myexpect_{q_i} \sqr{\loss_i} \\
      \implies &q_i \propto {q}_\jnt \frac{1} {\site_i^\old} \exp \rnd{ - \myang{ \vT(\vparam),  \natgrad \myexpect_{q_i}\sqr{\loss_i} } }\\ 
      \implies &\site_i \gets \site_i^\old \frac{q_i}{q_\jnt}\\
   \end{split}
   \label{eq:pvi_2ndline_derive}
\end{equation}
Next, we rewrite the third line in \cref{eq:pvi}. We start by noting that in the first iteration ${q}_\jnt \propto p_0$ and, because $s_i=1$, we can rewrite it as 
\[
   {q}_\jnt \gets \frac{1}{{\mathcal{Z}}_\jnt} p_0 \prod_{i=1}^t s_i = \frac{1}{{\mathcal{Z}}_\jnt} p_0 \prod_{i=1}^t \exp\rnd{- \losshat_{i} }
\]
Due to this property, in the second iteration, we can write the prior as 
\begin{equation}
   p_0 \propto \frac{{q}_\jnt}{\prod_{i=1}^t s_i^\old},
   \label{eq:prior_rewrite}
\end{equation}
where the $s_i$ from the first iteration became $s_i^\old$ in the second iteration. Thus, we can rewrite the second iteration as follows,
\begin{equation}
   \begin{split}
      &{q}_\jnt \gets \frac{1}{{\mathcal{Z}}_\jnt} {q}_\jnt \prod_{i=1}^t \frac{\site_i}{\site_i^\old} 
      = \frac{1}{{\mathcal{Z}}_\jnt} \rnd{\frac{ {q}_\jnt }{ \prod_{i=1}^t \site_i^\old } } \prod_{i=1}^t \site_i
      = \frac{1}{{\mathcal{Z}}_\jnt} p_0 \prod_{i=1}^t \exp\rnd{- \losshat_{i} } . 
   \end{split}
\end{equation}
This can be written again in the form of \cref{eq:prior_rewrite}. Therefore, by induction, the third line in the PVI update always matches the outer loop of the posterior correction algorithm.


%

We also briefly discuss the issues with the ``damping'' used in \citet{ashman2022partitioned}. They use the following damping scheme where $\rho$ is introduced to update $s_i$ but the update of $q_i$ does not have any $\rho$ (that is, it is set to 1),
\begin{equation}
   \begin{split}
      &q_i \gets \argmin_{q_i \in \mathcal{Q} } \,\, \myexpect_{q_i} \sqr{\loss_i + \log \site_i} +  \dkls{}{q_i}{{q}_\jnt} \\
      &\site_i^\old \gets \site_i,\,\, \text{ and } \,\, \site_i \gets \site_i^\old \rnd{ \frac{q_i}{ q_\jnt} }^{\highlight{\rho}}.
   \end{split}
\end{equation}
The issue with the update is that it creates a delay between the update of $q_i$ and $s_i$. Essentially, the site $s_i$ are not using the most fresh natural gradients $\dualparamhat_i \gets \natgrad \myexpect_{q_i}[\loss_i]$, instead we have a moving average:
\[
   \dualparamhat_i \gets (1-\rho) \dualparamhat_i + \rho \natgrad \myexpect_{q_i}[\loss_i].
\]
We do not know the consequences of this updating scheme on the overall convergence.

\bibliography{refs}

\end{document}